# Gatekeeping Algorithms with Human Ethical Bias
*The ethics of algorithms in archives, libraries and society*

dr.ir. Martijn van Otterlo
Vrije Universiteit Amsterdam, The Netherlands
http://martijnvanotterlo.nl
mail@martijnvanotterlo.nl

> *"We must bring together a collection of machines which simultaneously or sequentially can perform the following operations: (1) The transformation of sound into writing; (2) The reproduction of this writing in as many copies as are useful; (3) The creation of documents in such a way that each item of information has its own identity and, in its relationships with those items comprising any collection, can be retrieved as necessary; (4) A Classification number assigned to each item of information; the perforation of documents correlated with these numbers; (5) Automatic classification and filing of documents; (6) Automatic retrieval of documents for consultation and presented either direct to the enquirer or via machine enabling written additions to be made to them; (7) Mechanical manipulation at will of all the listed items of information in order to obtain new combinations of facts, new relationships of ideas, and new operations carried out with the help of numbers. The technology fulfilling these seven requirements would indeed be a mechanical, collective brain"* –-
> Paul Otlet in 1934, quoted by Rayward, W.Boyd (1992:91).

**Abstract**
In the age of algorithms, I focus on the question of how to ensure algorithms that will take over many of our familiar archival and library tasks, will behave according to human ethical norms that have evolved over many years. I start by characterizing physical archives in the context of related institutions such as libraries and museums. In this setting I analyze how ethical principles, in particular about access to information, have been formalized and communicated in the form of ethical codes, or: *codes of conducts*. After that I describe two main developments: digitalization, in which physical aspects of the world are turned into digital data, and algorithmization, in which intelligent computer programs turn this data into predictions and decisions. Both affect interactions that were once physical but now digital. In this new setting I survey and analyze the ethical aspects of algorithms and how they shape a vision on the future of archivists and librarians, in the form of algorithmic documentalists, or: *codementalists*. Finally I outline a general research strategy, called IntERMEeDIUM, to obtain algorithms that obey are human ethical values encoded in code of ethics.

## (1) Introduction[1]
In 2016 the smartphone *augmented reality* game Pokemon Go[2] became a hype. Looking through the phone creatures (the Pokemons) can be seen (projected onto the camera image) and virtually caught. Pokemon hunting became a very popular activity worldwide. April[3] 2017 still featured 5 million active players per day (and 65 million per month), the game was downloaded more than 650 million times, and revenues were high. The *transformative effects* on society were profound. For example, all players together had taken 144 billion steps[4] while playing, *because of the game*. That is, the game scatters Pokemons over particular areas on the globe, and players will have to go there, physically, to catch them. Some people have traveled thousands of kilometers[5]. Companies can even buy more Pokemons in some areas, for example to physically lure[6] potential customers to a shopping area. In the Netherlands, one particular beach area, which effectively became known as PokeDuin[7] (where Duin is Dutch for "dune") was flooded by players, annoying the local population.

---
1 All hyperlinks in the footnotes were last accessed October 2017.
2 http://www.pokemon.com/us/pokemon-video-games/pokemon-go/
3 http://expandedramblings.com/index.php/pokemon-go-statistics/
4 https://games.slashdot.org/story/16/10/11/2110219/pokemon-go-could-add-283-million-years-to-users-lives-says-study

Pokemon-Go shows us is that a relatively small simple *digital* game, or: an *algorithm*, can influence the physical world globally. Such effects are characteristic of the digital world, where everything is connected and scales of time and distance are profoundly changed. In this new, digital world, there is a need for *understanding* such changes, e.g. developing *algorithmic literacy*, and overall, rethinking the *ethics* of new and old patterns. For Pokemon Go ethical[8] questions arise such as *"is it ok to lure thousands of people to PokeDuin?"* and *"would it be good to increase the physical activity of people?"*, which in the latter case could help to fight obesity. Ethical reasoning is *practical* reasoning about "good" and "bad", while keeping an eye on "for whom" something is good or bad.

Our digital world causes many such ethical issues, and companies such as Google, Facebook, Twitter and Instagram influence this world. Digital innovations in big data, the internet-of-things, smartphones, profiling (van Otterlo, 2013) and the internet all connect "us" (the humans) with a growing body of data and knowledge. Intelligent algorithms are now mediating most of our interactions with this body of knowledge, but also interactions among ourselves and have unlimited means to manipulate us (van Otterlo, 2014a). More specifically, they monitor and determine the growth of the body of knowledge, they generate data themselves, and they determine the content and breadth of the view we (the humans) get of the world. Algorithms are thus becoming the ultimate *curators* and *gatekeepers* in our quest for knowledge (van Otterlo, 2016a). This has profound implications for how we deal with knowledge, what knowledge actually is, and who gets access to which kinds of knowledge. Archivists and librarians have long been the human gatekeepers, but increasingly so they are assisted, rivaled, or even replaced by intelligent, and even adaptive, algorithms. This raises, quite naturally, a flood of new, interesting ethical questions.

Access to lots of information has been the dream of many visionairs, especially in the last century. Joseph Licklider (1965) predicted more than fifty years ago that humans by the year 2000 would invest in a kind of *intermedium*[9] which would provide access to the so-called *procognitive net*, containing all knowledge. Paul Otlet envisioned various automated ways to do knowledge classification and retrieval, and laid the foundation for the modern internet with his Mundaneum and universal decimal classification. In 1945 Vannevar Bush introduced the "Memex", resembling Otlet's "Mundotheque" (introduced around the same time), a machine in the form of a regular desk that used microfilm as the storage medium for collections of text, and which could provide access to knowledge. Otlet's version was more related to H.G Wells "World Brain" in the sense that it focused on "networked" knowledge, and targeted film, photographs and radio in addition to text. Wells, building on ideas on information retrieval in his early "A Modern Utopia" from 1905, introduced his World Brain in 1939 in a series of lectures, as an idea to make the whole human memory accessible to every individual. More recently Wilfred Lancaster wrote (1982, pp33-34, quoting Schiller 1977): *"Ultimately, the signs point to a technology offering search capability at home or office terminals without the aid of librarian intermediaries who perform the searches."* All these, and many more pioneers (see Borner (2010) and Wright (2014) for extensive overviews), envisioned forms of technology that would connect each individual to "all" knowledge, in the form of some "world encyclopedia" and would make this knowledge retrievable by technology. In essence, our current world, with Google, Wikipedia, Twitter and smartphones, exhibits all that they were looking for.

The enthusiasm of these pioneers in "universally accessible" knowledge is echoed in today's Silicon Valley's technology push. Every day comes with new services, new technologies, new apps and new algorithms in the form of *artificial intelligence* (AI) (Nilsson, 2010). That each person on earth, in principle, has access to the world's knowledge through a smartphone was just a start. Soon, intelligent algorithms will take over many other tasks that have to do with the acquisition, organization, interpretation and use of the vast amount of information on the internet, and the ever growing heap of big data. What Silicon Valley and pioneers such as Otlet and Licklider also have in common, at least until very recently, is that their focus was on the *possibilities* of novel technologies, and not on possible (*unintended*) *consequences*. Despite some attention to privacy issues, only really starting in the age of databases (see Warner and Stone, 1970) often such topics were ignored. For example the book by Lancaster (1982) only mentions it once, and sets it aside as a minor issue. Today, however, privacy and security get more attention (e.g. hacking and spying), but as we have shown in the Pokemon Go example, digital technology, and especially intelligent algorithms, has much broader implications, especially when algorithms determine *access* to information (van Otterlo 2016a).

---

5 http://www.bild.de/lifestyle/2017/viral/dieser-mann-hat-ein-jahr-pokemon-durchgespielt-52281362.bild.html
6 http://www.dailymail.co.uk/news/article-3686473/Pokemon-Gos-digital-popularity-warping-real-life.html
7 https://tweakers.net/nieuws/115495/den-haag-schakelt-advocaat-in-om-minder-pokemon-af-te-dwingen.html
8 https://www.wired.com/2016/08/ethics-ar-pokemon-go/
9 In his words: *"a capital investment in their intellectual Cadillac"*.

Some recent ethical challenges with access are *fake news,* Pariser's (2011) *filter bubbles*, where algorithms reinforce people's biases, and *censorship*. As an example, Facebook's policy to allow or disallow particular content, essentially implementing a form of censorship[10], raises many ethical issues given their 2 billion user base. Recently some of it has been disclosed[11] but generally it is unclear who decides upon them. Facebook is also active in detecting utterances related to terrorism[12], Google aims to tackle fake news by classifying[13] news sources and marking them, effectively implementing a "soft" version of censorship, and Twitter targets[14] "hatespeech", thereby implementing language (and possibly thought) monitoring on the fly. These are just a couple of examples of *"access control"* by private entities with ethical consequences which were not foreseen by the earlier mentioned pioneers. Big technology companies are starting to recognize the ethical[15] issues, even causing Google to revive Wiener's[16] idea of an emergency button[17] to turn off autonomous systems.

Adding to the story on the tension between the "old", physical world and the "new" digital world in the context of *access,* we can look at Facebook's idea to *predict*[18] potential suicides. This is a delicate issue, since Facebook's "action repertoire" potentially includes prohibiting someone to access particular information, but also invoking a person's social network. Imagine getting a message about your friend saying *"Facebook's algorithm has analyzed your friend's posts for the last weeks and it has determined that with 0.63 probability he is likely to commit suicide, with medium confidence. Can you keep an eye on him?"* Here, Facebook's role as "guardian" of a social network dominates and it utilizes the data it has on users to provide such ethically-heavy services. In essence, it could make predictions of which users themselves are not even aware, for example people who don't know that they have a depression (Otterlo 2013). A completely different take on this was investigated by Juznic[19] et al. (2001). They used a "mystery shopper" tactic to investigate how *librarians* in *physical libraries* would react to "inappropriate" information requests by patrons, for example about necrophilia and photographs of dead people, but also very specific questions on "finding information on committing suicide". Interestingly, most librarians were not shocked at all by the request and treated it as a pure information request. Quoting the authors: *"Our conclusion was that the librarians in public libraries satisfied the need for information as much as they felt inclined to do so, and this was not affected by judgments about the ethical status of the required item of information."* One key aspect here is that whereas Facebook *judges* people and undertakes particular actions according to that judgment, the librarians in the latter research kept to their professional standards and focused on a bias-free, judgment-free information provision. Here we see a crucial difference between an old (human) and new (machine) information provider, with different views on what it means to provide access to information, but also a difference between opportunities offered in our data age.

---

10 http://fortune.com/2017/05/22/facebook-censorship-transparency/
11 https://www.theguardian.com/news/2017/may/21/revealed-facebook-internal-rulebook-sex-terrorism-violence
12 http://www.telegraph.co.uk/news/2017/06/16/facebook-using-artificial-intelligence-combat-terrorist-propaganda/
13 https://www.theguardian.com/technology/2017/apr/07/google-to-display-fact-checking-labels-to-show-if-news-is-true-or-false
14 https://www.forbes.com/sites/kalevleetaru/2017/02/17/how-twitters-new-censorship-tools-are-the-pandoras-box-moving-us-towards-the-end-of-free-speech/
15 https://www.wired.com/2016/09/google-facebook-microsoft-tackle-ethics-ai/
16 Wiener was, however, sceptical: *"Again and again I have heard the statement that learning machines cannot subject us to any new dangers, because we can turn them off when we feel like it. But can we? To turn a machine off effectively, we must be in possession of information as to whether the danger point has come. The mere fact that we have made the machine does not guarantee that we shall have the proper information to do this."* (N. Wiener (1948, 1961): Cybernetics, or control and communication in the animal and the machine).
17 http://www.dailymail.co.uk/sciencetech/article-3624671/Google-s-AI-team-developing-big-red-button-switch-systems-pose-threat.html
18 https://www.wired.com/2017/03/artificial-intelligence-learning-predict-prevent-suicide/
19 Thanks to Prof. Juznic for pointing me to this paper after my invited talk on ethics and libraries at the 9th International Conference on Qualitative and Quantitative Methods in Libraries (QQML) in June 2017 in Limerick, Ireland.

In this text I explore ethical implications of a gradual, but unstoppable shift from human-populated information-providing professions such as in libraries and archives towards situations where intelligent algorithms are quickly taking over many tasks of information *gathering*, *selecting*, *archiving*, *describing* and *providing access*. Before this *algorithmic turn* ethical issues were dealt with by humans, and the main question in this text is how to ensure that human values and norms are maintained in current and future algorithmic developments. Some of the material in this text is derived from the recent course[20] I developed on the ethics of algorithms, and recent work (e.g. van Otterlo 2013,2014,2016a,2016b).

**Outline and contributions**. First in the next section I discuss an archetypical archival practice and I connect it to similar information professions such as libraries and museums. I focus on the *physical* aspects and how they relate to *access* by users. In Section 3 I discuss the *ethical* issues related to access, as well as ways to *formalize* ethical guidelines in so-called *codes of conduct* which are sets of principles which *human* professionals should obey to be a *good, professional* archivist or librarian. Many codes exist, influenced by the profession, by norms and values in society at a particular time, and especially by advances in technology. Whereas Sections 2 and 3 mainly focus on concepts of the traditional, physical world, Section 4 moves to the digital world. It first defines a novel view on how *digitalization* affects virtually all aspects of life. Digitalization's core outcome is the so-called *big data* representing all our interactions, conversations, purchases, etc. I will discuss the digitalization of archives and libraries in particular, and review core ideas. A second concept defined in this section is *algorithmization*: the rise of intelligent *algorithms* that analyze big data to find patterns, e.g. to identify customer groups, or to decide upon important issues. Algorithmization also implies *replacing (core duties of) human professionals by algorithms.* Digitalization and algorithmization together represent the core mechanisms of a new view on huge transformations in society, and in professions such as archives and libraries. Section 5 then discusses old and new ethical challenges of algorithms, focused at a novel core taxonomy – based on agency and autonomy – which structures the space of algorithms effectively. I survey the novel field of *ethics of algorithms* and connect to core operations of archives and libraries. In Section 6 I develop the novel IntERMeDIUM research strategy to ensure future algorithmic information professionals, which I will call *codementalists*, will behave according to human values contained in codes of ethics, and Section 7 concludes the text.

**(2) Gatekeepers in Physical Settings**
Here I first deal with the *physical* world of archives and libraries. The next section describes ethical aspects, i.e. how to arrange for "proper" or "fair" access.

**(2.1) The Physical-Analog World**
As explained in the introduction, this text is about the *transformation* of archival and library practices, and society in general, to the digital age, and its consequences. To describe a transformation, I first need to introduce the starting point: the *physical-analog* world. Another reason for that is that it is familiar to humans: we have been raised with the perception and manipulation of our surrounding physical world. Abstract mathematical ideas and complex technical procedures are often, and for many people, difficult to grasp. However, if we can have a look at something, and play around with objects, we often understand complex items like machines, (board) games and utensils quickly. In previous work (van Otterlo, 2016a) I utilized *"libraryness"*, which exploits our familiarity with physical-analog library buildings, books and processes. This concept can induce a *metaphor* for how *intelligent algorithms* can be seen as a *librarian* in a giant *digital library,* providing an intuitive setting to understand complex intelligent software systems.

---

[20] http://martijnvanotterlo.nl/teaching.html

Physical worlds can also provide understanding of other physical worlds using analogies. Peekhaus describes[21] so-called "seed libraries" in the United States, which are public libraries that maintain various seeds of plants and vegetables, instead of books. Each patron can take some seeds, plant them, grow a crop and later extract new seeds and return these later. This allows a community to form based on the sharing and growing of vegetables. This novel concept sounds interesting, but only by making an analogy with a known situation – in this case a traditional library – one gets to see its novelty. Namely, in terms of traditional libraries, it amounts to taking a book, keeping it with you for one or two years, in the mean time using it to write a new book (or several new books) yourself, burning the book you borrowed, and giving one or several of your new books to the library. Now that sounds interesting indeed!

Physical worlds require physical actions, dependent on time and space, which induce *constraints* on the types of things that can be done. For example, rearranging 25.000 physical books in a personal library amounts to serious physical work whereas the same *operations* on an e-book library would be much easier. Especially for books and libraries, physical aspects are important. These include architectural aspects, location, functional spaces, social spaces (e.g a coffee room), furniture and shelves (Edwards, 2009). For hundreds of years, until very recently, books and libraries were always physical: pages, paper, bricks, shelves, etc. The interaction with books is *embodied,* and thus highly related to physical, sensory experiences in terms of taste, sound and smell, but also the haptics of reading and writing, note-taking, and the emotional value of physical books. Reading, writing and handling of books relies on sequential access to pages, on the visual nature of page design, on the embodied aspects of handwriting, and more. Technology is redefining what it means to *read* (Baron, 2015). Piper (2012): *"Things ask us to do certain things with them. Things are not unconditional. […] They (books, mvo) still shape our access to what we read and how we construct our mental universes through them."*

Overall, the physical nature of archives and libraries is important and intuitive. First, our brains have evolved dealing with the physicality of the world and have shaped our understanding of, and interaction with, the world. Second, the physical world also shapes our actions: it puts restrictions on the things we can do. Third, many aspects of libraries and archives have been physical for most of the time in our human existence, and only very recently things are rapidly changing, transformed by technology, and it is to be seen what we gain and lose (Baron 2015, Piper 2012, van Otterlo 2016, and the rest of this text).

**(2.2) Museums, Archives and Libraries**
There are several institutions in our society that collect, maintain, and provide access to, *information*. These institutions are known as *museums*, *libraries* and *archives* (Kirchhoff et al., 2008): physical places where "interesting" items are stored, or *preserved*, for society as a whole, or more private groups such as companies. Typical professions in each of them are *historians*, *librarians*, and *archivists*. Most people will know some differences between these three kinds of "memory institutions"[22].

Most cities in the world features many *museums* which maintain and display a (themed) *collection* of items. Items in a museum collection are not typically *used* or *taken by visitors*, but they can be looked at and the museum provides additional information through signs, tours and other means. The main purpose of a museum is to provide a place where people can learn about a particular topic, or merely enjoy the collection, such as with *art*. Many museums ask for an *entrance* fee but do not typically restrict access for particular

---

21 See *"An exploratory study of seed libraries in the United States"* (abstract presented at QQML2016). http://www.isast.org/images/e-_Book_of_Abstracts_final_2016_v9.pdf (p60)
22 Kirchhoff et al. (2008, p252) cites Lorcan Dempsey (2000) as follows: *"Archives, libraries and museums are memory institutions: they organize the European cultural and intellectual record. Their collections contain the memory of peoples, communities, institutions and individuals, the scientific and cultural heritage, and the products throughout time of our imagination, craft and learning. They join us to our ancestors and are our legacy to future generations. They are used by the child, the scholar, and the citizen, by the business person, the tourist and the learner. These in turn are creating the heritage of our future. Memory institutions contribute directly and indirectly to prosperity through support for learning, commerce, tourism, and personal fullfilment."*

groups.

A *library* (Manguel 2006; Palfrey 2015; Crawford 2015), equally familiar to most people, is something very different. Libraries come in two main kinds: public libraries and academic libraries. A public library is a *public place*, centered around a usually large *book collection* from which users – or in library language, *patrons* – can take items and use them there or at home for a while. Public libraries also provide many other things[23], such as working spaces for students, internet access for the public, and typically they organize activities to aid a local community. Academic libraries are different, in that they are mainly supporting a university, which determines their collection but also its use. The physical place of the university library functions as a kind of public space (Aabø and Audunson, 2012) too, typically filled by students with laptops, but the collection is often predominantly digital. Modern universities are much focused on *data and digital services* and *digital access* to resources rendering the librarian more of a *data broker*. Both types of libraries have in common that their physical collections can be browsed by visitors who can take items and use them, without access restrictions. Often libraries have so-called *special collections* for which there are restrictions on who can access, use, take or even touch specific items. For example, a library may feature a collection of *incunables* – books printed before 1501 in Europe – which are very delicate. Access to such books may be required since the physical, original copy may show characteristics which get lost in digital reproductions, but special permission is then required to protect the book.

*Archives*, the third type of memory institution, are maybe the lesser known of the three. Archives are similar to the special collections with forms of restricted access, and similar to museums which feature *any* type of items, not just books. Morris (2009:4): *"Archives are records, regardless of format, created or received by a person or organization during the conduct of affairs and preserved because they contain information of continuing value."* In addition, *archive* can refer to the physical building or to the *agency* or *program* responsible. There are many types[24] of archives, typically preserving materials for governments, but also individuals or private corporations, for example to keep historical records to serve the needs of company staff members and to advance business goals. In an organization, "archive" is usually the "non-current records". Examples of records are policies and procedures, meeting minutes, annual reports, correspondence, photographs and audiovisual materials. In archives for individuals (or families) the records are called "manuscripts", "personal papers" or "family papers". Archives are often regulated by law; for example companies are legally required to keep their financial records for reporting purposes.

An interesting *physical* archive is the *Lego vault*[25] at Billund[26] in Denmark where the headquarters of the Lego company[27] is. This vault contains all Lego sets ever being put on the consumer market, all present as the well-known (unopened!) carton boxes, on movable archival shelves, ordered by theme, by year and so on. These sets physically carry "information of continuing value", since they represent a history[28] of the core outcome of the company's innovative[29] activities: Lego sets. Access is highly restricted and maintenance is governed by the Lego vault archivist. Physical constraints on the space are severe and new space needs to be found to accommodate all new[30] sets. The Lego vault is an exciting phenomenon, but it is only one part of a larger company archive that also consists of letters, financial statements, internal and external communications, advertising materials and much more. One may wonder why my own Lego collection, which is neatly organized into hundreds of categories, or a Lego factory[31] warehouse, where millions of pieces are stored in a huge inventory, are Lego *archives* too. An important aspect of archives is that *context* is everything: they reflect the lives of or activities of the record creator, and records are created "during a conduct of affairs". Inventories and personal collections typically are not archives in the formal sense, since they do not contain "information of continuing value". Of course, the bricks have value, but are not preserved for long-term reference, nor have historical value in that context.

---

23 Public libraries in the Netherlands have a rich history of reorganization (see Keizer, 2017).
24 https://www2.archivists.org/usingarchives/typesofarchives
25 https://www.youtube.com/watch?v=EgFvzpB6BsQ by the makers of the website beyond-the-brick.
26 http://www.minibillund.nl/ is a Dutch store and Lego museum (or archive) in Wagenberg.
27 http://www.lego.com
28 http://www.brickshelf.com/gallery/hmillington/Misc/LEGOLife/legolife.jpg in which the administrator of the well-known Lego resource site http://brickset.com visits the Lego vault in Billund.
29 http://www.wired.co.uk/article/building-success
30 2016 features more than 800 sets (https://brickset.com/sets/year-2016)
31 https://www.theguardian.com/lifeandstyle/gallery/2014/aug/29/inside-the-lego-factory-in-billund-denmark-in-pictures

Museums, archives and libraries are distinct, yet also very similar: they all professionally maintain an *information collection* and take on the role of *gatekeeper* to that information. Some argue that they will *converge* to a unified institute, especially in digital developments (Kirchhoff et al., 2008; Zastrow, 2013). *"The avalanche of born-digital records, digitization projects, digital curation in special collections and new fields such as digital humanities has finally brought about an intersection of these various disciplines."* (Zastrow 2013:16). Others, such as Clement et al. (2013:113) state similar ideas (quoting Jerome McGann): *"Libraries and museums – let's call them archives."*, and quoting Marlene Manoff (on page 112): *"as libraries, museums and archives increasingly make their materials online in formats that include sound, images and multimedia, as well as text, it no longer makes sense to distinguish them on the basis of the objects they collect."* Especially in North-America, such convergence can also be seen in the rise of the "i-schools[32]" which promote an interdisciplinary approach for information management.

**(2.3) Core Archival Processes**
To make things more tangible, let us now look at some of the *core[33] processes* when working with archives, and mark some[34,35] differences with the more familiar libraries.

A first set of operations in the archive[36] is about the **selection** of materials, i.e. what gets into an archive in the first place. In a typical archive, records come from an organisation as outcome or part of company processes in the **acquisition** phase, whereas for libraries new entries come mainly from new, published books. Continuously archivists evaluate records for permanent value, appraising them for *disposal* or **retention**. They serve as historians for the organization that creates the records, preserve and protect the records for ongoing use, and ensure legal compliance regarding retention periods and accessibility of records. Through **appraisal**, archivists determine which records belong in the archives, based on their long-term administrative, legal, fiscal, and research value. Through **acquisition**, archives obtain those records which meet the appraisal criteria. Through **accessioning**, the archivist takes physical control of records by transferring them to the archives repository and begins establishing intellectual control. Archives can hold both published and unpublished materials, in any format. Destroyed records can never be recovered[37], since archives often contain unique, physical items. For librarians this is less of a problem since they typically deal with (bulk-)published materials which can be reproduced when needed. In the Lego vault, acquisition would amount to getting all new Lego, and appraisal would, in this case, only throw away duplicate sets.

A second set of processes involves the **maintenance** of the archive. Simply put, it deals with all the processes of placing, ordering, and physically keeping the records in a physical space. **Provenance** is a fundamental principle of archives, referring to the individual, family, or organization that created or received the items in a collection. The principle of provenance (or: the *respect des fonds*) dictates that records of different origins (provenance) be kept separate to preserve their *context*. In the Lego vault, sets from particular lines, or years, get their own place on the shelves. The **order[38],** or **physical arrangement** of the records in the physical archive is important[39], and puts severe constraints on how to *layout* the collection of

---

32 https://en.wikipedia.org/wiki/Information_school
33 https://www2.archivists.org/node/14804
34 https://www.quora.com/How-would-you-explain-the-difference-between-a-librarian-and-an-archivist
35 https://www2.archivists.org/usingarchives/whatarearchives
36 https://www2.archivists.org/about-archives
37 Piper (2012:20): *"Scholars of the future will no doubt troll libraries to locate 'lost' print editions of undigitized texts, just like their print predecessors scoured libraries for lost manuscripts."*
38 One idea to make the ordering of books easier is to have strict maximum number of books in a library. Perec (1985:148-155) sets it at 361, but he also notes that (p152) *"Disorder in a library is not serious in itself; it ranks with 'Which drawer did I put my socks in?'. We always think we shall know instinctively where we have put such and such a book. And even if we don't know, it will never be difficult to go rapidly along all the shelves."*. While I think, or maybe even know, this holds for small, personal libraries, it is generally not true.
39 *Habeat Librarius et registrum omnium librorum ordinatum secundum facultates et auctores, reponeatque eos separatim et ordinate cum signaturis per scripturam* records over the *physical space.* Even for libraries, with only books as "records", ordering is not simple[40]. **Preservation** even gives records "new life" after the "death" of their original medium by migration of records from one medium to another, e.g., photocopying to alkaline-buffered paper, microfilming or digitizing records, or periodic re-copying of film-based or digital records. The **arrangement** of items in libraries is predetermined based on subject classification (e.g. Dewey decimal), but there is still a challenge to map this knowledge classification scheme onto the physical space (i.e. floor plan of the library). In archives it is determined by provenance, original order and other types determined *organically* by the archivist based on deep knowledge about the originating organisation. The final arrangement of materials will usually be alphabetical or chronological within record groups or series, showing the hierarchical relationship of each fond (creating office or individual) to the institution's other fonds. Overall, whereas in libraries the ordering of items is quite standard, the one in archives has to be determined by the archivist.

The third set of operations aids in *finding* things. The **description**[41] of the contents of the archive in a **catalogue** is important for the actual **use** of the archive, which also requires a **reference system** detailing *where* items from the catalog can be found in the physical space. What is characteristic for archives (and not for libraries) is that content is often only described at an *aggregate* level: for example there could be a box containing tax records of year 2016 with no detailed description of the individual items in that box. Librarians are used to work with secondary resources (books, serials) which can be described by call number and cataloging information inherent in the book itself or available through bibliographic databases such as WorldCat. They can often download the bibliographic information and therefore the cataloging (in the original sense) is not much part of the daily work. Archival materials are acquired without descriptive call numbers or assigned object headings and titles. Archivists create an accession record – noting the records' date, title, bulk, condition, transferring office or donor, conservation needs, and access restrictions – when records come into the archives. Archives are difficult to catalog in traditional library systems; they typically feature unique, rare, valuable, frequently agile, and difficult-to-decipher materials. Full record description is one of the most complex and challenging archival tasks. It involves knowing what the item is, what it means in the archive as a whole, and knowing in advance how people will/may search for it.

Summarizing, physical archival processes concern three things: 1) **selecting** (and discarding) which items belong in the archive, 2) **ordering** and physically arranging the records in a physical layout, and 3) describing and **documenting** which records exist and where they are. Compared to libraries, archives are more idiosyncratic and require more creativity and more work mainly because of the variety of materials and because orderings and descriptions sometimes need to grow organically, locally, as opposed to library systems which are more standard, dealing solely with books, and of which catalogues can be shared among institutions. The trend in libraries is to let users interact more and more without any help of a librarian[42] whereas in archives the archivist is still the main contact person, or the main gatekeeper, to the archive.

**(2.4) Access**
Physical access to archives and libraries has always appealed to our imagination, in *fiction*, *poetry* and *film* (Crawford 2015). Exciting stories like[43] *Indiana Jones* revolve around the idea of finding a lost archive and

---

*applicatis* Eco (1980, p84) *"The librarian requires a catalogue of all books, ordered by sciences and authors, and he needs to place all books individually and orderly, accompanied by written marks."* (the author's translation).

40 Perec (1985:152-153), Edwards (2009:106-111) and Petroski (in the appendix "Order, order") list many options for shelving books and ordering them by their characteristics.

41 https://www2.archivists.org/usingarchives/appendix

42 See also the discussion on "retail libraries" in (van Otterlo, 2016b).

43 See also the three Librarian movies (2004) (http://www.imdb.com/title/tt0412915/) and (2006) (http://www.imdb.com/title/tt0455596/) and (2008) (http://www.imdb.com/title/tt1146438/) about the librarian of an imaginary archive containing things like Noah's Ark and Pandora's Box. Furthermore the Indiana Jones movie series (https://en.wikipedia.org/wiki/Indiana_Jones) features many archeological treasure hunts. Other popular hunts in archives and libraries can be found in "National treasure" (http://www.imdb.com/title/tt0368891/) and its sequel "Book of secrets" (http://www.imdb.com/title/tt0465234/), and of course Name of the Rose (http://www.imdb.com/title/tt0091605/). Yet another series of movies centered around the idea of libraries and archives are: The Da Vinci Code (http://www.imdb.com/title/tt0382625/), Angels and Demons retrieving a valuable item. The nicest example of such a *physical hunt* for a book appears in Umberto Eco's (1980) *Name of the Rose*, in which the narrator, Adso, assists William of Baskerville in a murder investigation in a medieval Abbey. At one point[44], they find the entrance to the secret library in the forbidden tower of the abbey, by pressing the eyes of a skull on a tomb stone. Adso and William enter the library, impressed by all the books they find, and go look for the only remaining copy of Aristotle's *Second Book of Poetics on Comedy*, which seems somehow related to the murders. William: "*Do you realise, we are in one of the greatest libraries in the whole world!*" William thinks that the reason why the library was kept hidden was the difficult issue of the content of the library, advanced knowledge coming from basically pagan thinkers, and no easy match with Christianity. Adso says: "*How are we going to find the books we are looking for?*" William: "*In time!*" Seeing all these books impresses the men, but they also feel that they should be accessible to everyone. William: "*No one should be forbidden to consult these books!*" But, the sheer amount of books makes them prepare for search. William: "*How many more rooms? How many more books?*" In the end, William and Adso go even deeper into the labyrinth of the library and finally find Jorge von Burgos, the blind[45] librarian, who beat them getting into this part of the library. (Eco, 1988:421)*: "if anyone wanted to know the location of an ancient, forgotten book, he did not ask Malachi, but Jorge. Malachi kept the catalogue and went up into the library, but Jorge knew what each title meant."* Jorge: "*You have discovered much about the abbey, but the short route to the library is not among them.*" By finding Jorge, William and Adso finally find Aristotle's book, which is on humour, a topic deemed unfit for the library by Jorge. The book has poisonous pages and is the murder weapon. The core of this story is about *access* to knowledge in the physical manifestation of a library or an archive. The complex labyrinth, the evil librarian Jorge, and the absence of a catalogue system made for heavy, deliberate *physical access control mechanisms* that tried to prevent Adso and William from finding the book (or even the library itself).

In another text[46] Eco describes a (second) hostile library, now to ironically point out 18 guidelines for how (not) to run a library properly. Some of these include: books should not carry any information about the category they belong to, there should be ridiculous restrictions on what can be borrowed and when, the information desk should be inaccessible, and you should not be able to find the book you were reading today, tomorrow again. Overall, (guideline H): *"The librarian should consider the reader as an enemy, a slacker (otherwise he would be working) and a potential thief."* Both Eco's libraries instate mostly physical barriers to *access*. Restricting access in libraries and archives can have many reasons. Since materials in archival collections are often unique, the archivists in charge of caring for those materials strive to preserve them for use today, and for future generations of researchers. Archives have specific guidelines on how people may use collections to protect the materials from physical damage and theft, keeping them and their content accessible for posterity. In general, a difference between libraries and archives is that plain **access** usually is more restricted for the latter. Libraries feature *open stacks* from which users can take items themselves; archives feature mainly *closed stacks* and prevent unrestricted circulation of records. Most archives will also limit photocopying, photographing, or scanning of archival documents to ensure preservation and security and respect copyright law.

Abstracting a bit away from physical book hunts and hostile librarians, access to archives can be seen in two different ways. One, the one which we have seen most so far, is related to *control* and *security*. It is targeted at restricting and controlling the (type of) access a user has to particular records, to the catalogue, or even to the knowledge of where (parts of) the archive is. Some archives also require users to provide identification and register the records requested. These data could be permanently retained in case of theft or misuse of items and for the archives' statistics. Maintaining statistics of use and records of visits is important for reports

---

(http://www.imdb.com/title/tt0808151/), and Inferno (http://www.imdb.com/title/tt3062096/). A good advertisement for archives comes from the Millennium Series (https://en.wikipedia.org/wiki/Millennium_(novel_series))) where the mystery is solved after many hours of labour in the company archives.
44 In 1986 a movie based on the book came out, with Sean Connery in the role of William of Baskerville; here I base some of the text on both the movie and the book.
45 This is an obvious reference to Jorge Luis Borges, the blind librarian, see (Manguel, 2006:248-249). Borges was the blind librarian of the National Library of Argentina for years (the third blind librarian in there). Borges acknowledged the irony of God giving him both all these books, as well as darkness.
46 In "De Bibliotheca" (1981); Dutch translation ("De bibliotheek") by Martine Vosmaer (1988:12-16).

and publicity, as well as for evaluations and planning future policies. Controlling access makes the archivist literally a *gatekeeper*[47] who (physically) determines which parts of the archive the user is allowed to see and use. The first type of access deals with why users *should not* get (full) access to the archive. The second type of access control relates more to the role archivists (and librarians) take on professionally: a *service-oriented* role in connecting users with the contents of the archive. For example, a user could go to an archivist (or librarian) with a particular query, and the archivist would then help the user with finding the right parts of the archive. Furthermore, archivists could also play an *active* role in pointing users to other sources that could be of interest. This relates to the broader topic of promotion: to let the public know your archive exists and to promote its use. The "statistics of use" which may be gathered may also be used for benefit for the user, something that was recognized early on in the *mechanisation of libraries*; quoting Don. R. Swanson in (Markuson, 1963:13-14): *"Any user could then readily recover anything which he (or any other person he names) has used previously. One could ask for "the red book which I checked out last week" or perhaps 'the set of 15 books on automation which I used last fall'. An entire 'private demand library' could be rapidly constructable".* A big difference between librarians and archivists is that since the latter are much more involved in the selection, classification, description and arrangement tasks, users will stay more dependent on them, especially for the second kind of access. Overall, all actions in the archive (or library) have an influence on its use: collections, access policies and the archivist's service provision determine whether the user finds what he or she needs.

Much of this discussion so far, especially the second form of access control, comes back to the visionary information scientists introduced in the first section: Otlet, Wells, Licklider, Lancaster, Garfield and others. Their ideas about access traced back to the famous library of Alexandria[48],[49]. Over two millennia ago, the Ptolemites attempted to create a complete corpus of Greek literature (and some other languages). The library that held this collection, the famous library of Alexandria, can possibly be considered the first *universal library* (White 2008; Nerdinger 2011b). The medium in Alexandria was scrolls, later this became books, and nowadays we have digital records in some "cloud". Universal *access* means: being able to access *all* sources of knowledge available. The term "universal library" originates from Conrad Gessner (16th century, see Wright 2014) who gathered lots of sources, ranging from personal notes to parts of books, and rearranged them into new books among which the "Bibliotheca Universalis".

If universal access is desired, then in the purely physical world this means that physical copies of all books need to be assembled at the same geographical location. Technology and new media can help to loosen this constraint, something that has been realized in library mechanisation (Markuson, 1963), the electronic library (Lancaster, 1982) and the digital age (the rest of this text); Torres-Vargas (2005:156): *"In our time, due to the appearance of modern information and communications technologies, universal access has begun to seem like a realisable goal"*. She contrasts Otlet[50] (a social scientist and founder[51] of thinking about feasible universal access, and documentation and information science) and Wells (a writer in a scientific context), and shows that where Otlet aimed at a *centralization* of everything in his Mundaneum (copies of all the books in the world, and a universal bibliography), Well's (1937) idea of a a permanent World Encyclopedia was based on a *network* of connected institutions, more related to Licklider's (1965) ideas. Otlet's vast collection would be represented in the RBU (*Repertoire Bibliographique Universel*) that would allow identification but not access. The creation of an RBU reflects the age-old dream of recording everything that has ever been published anywhere in the world. Otlet's system was designed for documents, whereas Wells' was designed for information intended for the researcher. Wells' objective was not centralisation, but the extensive distribution of information. Torres-Vargas (2005, 156): *"Wells claimed that the encyclopedia should be organised in such a way that it did not depend on a single location, but was a kind of network. It could be mentally centralised, but not physically, and could therefore be duplicated."* In addition, his ideas are not based on physical possession of documents, and it can therefore be observed that the World Brain is closely related to the universal access conceived today with the use of networks and other information technologies. A related idea by Bush, the "Memex"[52], brings in (just like Otlet's Mundotheque) a physical *desk* from which a user can obtain access to books stored on microfilms. However, it was envisioned as a relatively stand-alone system, whereas Otlet's Mundotheque was a more networked idea, just like Wells'.

---

47 See Bozdag (2013) for additional pointers to the literature on *gatekeepers*, a phenomenon studied extensively in Media studies and journalism, see also (Granka, 2010).
48 https://en.wikipedia.org/wiki/Library_of_Alexandria
49 Gooding and Terras (2017) survey the typical use of *Babel* and *Alexandria* as metaphors for the digital library.
50 Otlet created the Mundaneum, a physical manifestation of the idea of universal libraries and universal access. Otlet is a fascinating figure in documentation (information science) with key contributions such as the Mundaneum, the universal Dewey decimal system (UDC), the Universal Book (similar to Gessner's work) and other things such as the universal atlas, the World Palace, and the utopian dream of a World City and ideas on the League of Nations. Pointers for further information: https://en.wikipedia.org/wiki/Mundaneum https://en.wikipedia.org/wiki/Paul_Otlet and Wright (2014), Borner (2010), Boyd Rayward (1992, 2010).
51 Torres-Vargas (2008:159): *"In addition to his efforts to create bibliographical institutions like those mentioned above, Otlet also wrote such important works as his Traité de Documentation, the masterpiece he published in 1934, which has been described by Rayward (1990) as a masterly exercise in synthesis, the first modern, systematic discussion of the general problems of organising information and one of the first manuals of information science. This essay formally coined the term 'documentation' to refer to the discipline involving the storage and retrieval of information."*

One conclusion of this section on physical archives (and libraries) must be that archivists and librarians have a powerful role as a gatekeeper to information. More than the librarian is the archivist involved in the *selection, arrangement* and *description* of the materials but (because of that) also in the *access* to the materials. In physical archives the user is dependent on the archivist to find the things related to his information needs. With such great powers comes great responsibility. Therefore, the in the next section I will look at the ethical dimensions.

**(3) Ethical Aspects of Information Access**
The previous section has dealt with *physical* access to *physical* archives and libraries by *physical* users. I have emphasized the physical aspects to make a clear distinction with our modern, digital world. Access in the physical world has a clear meaning, and also obstacles to that same access can be visualized in the same way. In Eco's library access can be controlled by maze-like buildings, by hidden doors, and by evil librarians making the physical search difficult or dangerous. The conditions under which such *access control* is enforced are deliberately chosen. Debating whether such choices are "good" or "bad" is the domain of *ethics*. In this section I will illustrate several specific ethical aspects in the archival profession, which shares many features with the library profession.

**(3.1) Ethics**
Eco's librarian Jorge's *moral values* (believing that laughter was sinful) made it appropriate to poison the book such that anyone who would read it would die a quick but painful death. Taking practical action based on moral values is the domain of *ethics* (Laudon 1995; Baggini and Fosl 2007; Baase 2013; Kizza 2013). According to Kizza (2013:3) *Morality* is *"a set of rules for right conduct, a system used to modify and regulate our behavior."* It naturally has close ties to *law* since when a society deems certain moral values to be important, it can formalize such values in a law and set behavior that will uphold those values as a *norm*. Ethics typically is concerned with analysis of such norm-setting processes. Classic ethical questions are: *"should we clone humans?"*, *"is it sometimes allowed to kill people?"* and *"should we provide a base income in case robots take over most jobs?"*.

As Laudon defines it (1995:34): *"Ethics is about the decision making and actions of free human beings. When faced with alternative courses of action or alternative goals to pursue, ethics helps us to make the correct decision. Ethics helps provide answers to questions like 'What should I do? What should we do? What goals should we pursue? What laws should we have? What collective behavior should we all pursue?'*

---

[52] (Loftus, 1956) *"Describes in detail a mechanized private file and library which is referred to as "Memex". In order that this mythical device might have a name, the author coined one at random. A "Memex" is a device in which the research worker may store all his books, records and communications. Information stored in this mythical machine can be consulted with extraordinary speed and flexibility. Since the mind operates on the principle of association of ideas, the author believs that information should be stored mechanically in the same manner. The "Memex" is so designed to make this possible."* Note that there are different versions about the origins of the name. The Dutch Wiki page refers to "Memory Extender" whereas the US Wiki page talks about "Memory Index".

*Ethics is concerned with practical decision making and human behavior in the broadest context."* To make it even simpler (Laudon 1995:34): *"Ethics is, above all, about what is good and what is evil, and how we come to make such judgments".* I would summarize it as: if there are options what to do, then ethics is concerned with practical reasoning about "good" and "bad" actions. Eco's librarian had multiple options, including the option to proactively give the book to Adso and William, and so we can compare his chosen option to this one and *evaluate* which one is better. Of course, this is according to *our* moral values. This already shows an important dimension: *for whom* is something good or bad, and *by who's standards*?

The field of ethics has many subfields dealing with subtle and specialized issues. Many of them are now applied or updated to computing and digital technology (Kizza 2013; Baase 2013). Here it suffices to look at four main *schools* of ethics (see Laudon 1995) induced from two dimensions: 1) *rules vs. consequences* and 2) *individual (micro-level) vs. collective (macro-level)*. The first dimension looks at where the moral values come from. For *rule-based* theories, the values come from outside, for example a religion. Rules simply say *"one must do X when Y"* or *"one is forbidden to do W when Z"*. A rule-based decision in traffic would be not to drive through a red light just *because* the rule say you are not allowed to do that. A *consequentialist* on the other hand, would look at the *actual consequences* and would, for example at night when the streets are completely empty, drive through a red light when appropriate. The second dimension looks at "for whom" the values or consequences have an influence (and "where" the moral authority is located). *Individual-based* theories look purely from the standpoint of an individual, whereas *collective-based* theories will look at (all) members of a community. The combination of the 2 dimensions results in the four main ethical schools of thought. The *rule-based collective ethics* is well-known, with typical scholars such as Socrates, Plato and Kant. The latter introduced the *"reductio ad absurdum"* to test whether something is ethical: *"shall I throw my litter on the ground? No, because if everybody would do that, it would become a mess!"* Individual rule-based ethics is typically associated with intuitions and religious belief whereas *individual consequentialist ethics* is more ego-centric and has more in common with neo-liberal, economic theories such as by Adam (*"the invisible hand"*) Smith and Ayn Rand. Making decisions according to this school amounts to optimizing self-interest. The fourth, and probably most intuitive and widely used, school holds the *collective consequentialists*. Here, ethical decisions are made according to *"the best balance between all interests of all stakeholders in decision making situation",* implementing John Stuart Mill's *utilitarianism*, which (Laudon 1995): *"advised us to take the actions that provided the greatest pleasure for the greatest number"*. Utilitarianism looks at possible consequences of a decision and values the *outcomes*. By comparing all outcomes for all stakeholders (and possibly weighting), one can optimize what is best[53] for all. Many real-life compromises are based on utilitarian principles. A main difference between rule-based and consequentialist ethics is that in the former the behavior is directly prescribed, whereas in the latter *reasoning* is needed to *calculate* the best decision based on the (predicted or perceived) consequences.

As an example, the Dutch national security agency has announced that it is trying to hack[54] the widely used Whatsapp-messenger service, in order to listen into conversations of (suspected) criminals and terrorists. Thus, the consequences of this would be that society becomes more safe (*positive consequence for the society at large)*, and terrorists would go to jail (*negative consequence for a very small group)*. However, it would also be a serious breach of *privacy* for all normal citizens who use Whatsapp (*negative consequence for a large subpopulation*). Typically politicians now need to weigh the various consequences, the amount of people per consequence, and the relative importance of them (e.g. negative consequences for terrorists are valued very low). Often such issues are resolved through *law*. For example, there are privacy laws protecting citizens from such hack-plans, but the same law can also make exceptions when national security is at stake.

---

53 One very good way to look at ethical dilemmas is to cast them into a scenario (Wright et al. 2014). That is, even though scientific and governmental reports can help to assess the impact and dangers of developments – such as new technologies – it often helps to also consider a more narrative account in the form of a scenario or fiction novel. Famous books such as Brave New World by Aldous Huxley and 1984 by George Orwell have caused generations to think deeply about technology, surveillance and the organization of society. Eco's story about the library too is a good scenario for archival ethics, and it can also be used to understand contemporary developments in *algorithms* (van Otterlo, 2016a). With a scenario, other ways become available to look at the degree of impact, the degree of uncertainty and new points of view, for a particular development. As Wright at al. say (2014:325): *"We subscribe to the dictum of scenario guru Peter Schwartz who defines scenarios as "a tool for ordering one's perceptions about alternative future environments in which one's decisions might be played out… Concretely, they resemble a set of stories."* By identifying stakeholders, applications, risks, drivers and technologies, one can try to imagine the various consequences of new developments.
54 http://www.telegraaf.nl/binnenland/27257015/__AIVD_kraakt_WhatsApp__.html

Note that ethical frameworks are not the definite answer to many problems; they only define a particular way of balancing moral values and behavior. In addition, there is something called *bounded ethical rationality* (Laudon 1995): it is impossible to assume that all necessary information is available, or can even be overseen, when making decisions. Also the tension between the individual and the collective is a factor: how to behave morally in evil societies?

**(3.2) Ethics of Archival Practices**

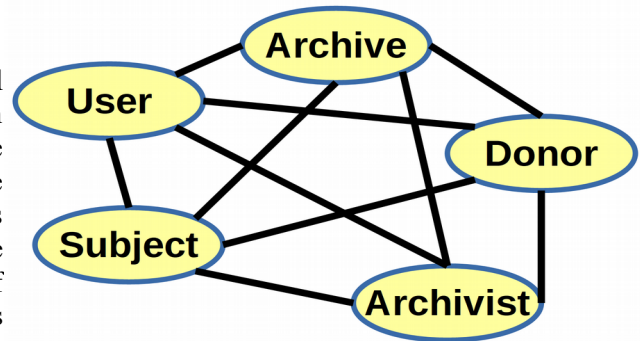

Now let us turn to how ethics applies to archival practices. Any relation between any of the five main components of archival practices – the archive, the user, the archivist, the donor organisation and the people occuring as a subject in the archive – is subject to choices, and ethical decisions. As we have seen in the previous paragraph, ethical analysis of each of these relations may be done from various viewpoints (one of the five individuals or any sub-collective) and also in terms of consequences versus rules. One possible rule could be that the archivist should provide anyone with access to the archive, but what if there is an expensive fragile vase? That requires some physical and ethical balance. The archivist plays a pivotal role in all ethical issues, but is sometimes overridden by (ego-centric) stakeholders such as the archive's donor organization and users.

Ethical challenges have always existed. A typical object of ethical study in this domain is *privacy* (Garoogian 1991, Svensson et al. 2016, van Otterlo 2013). Several other ethical dilemmas are about *access* (see next section) but plenty others arise between archive stakeholders. For example, Preisig (2014) mentions that unlimited freedom of expression collides with protection from defamation: archives may contain information that, when published freely, could cause harm to individuals (rendering a conflict with the owner or the subject of the archival matter). Ferguson et al. (2016) introduce a long list of 86 real-world ethical cases and cluster them by *dilemma*. Similar to Preisig et al's (2014) dilemma is the *"privacy versus potential harm to individuals"* but also included are *"privacy versus organisational ethos or requirements"* – where obligations to core customers were in conflict with the organisational interests, for example when a professor requests reading records of a student suspected of plagiarism – and *"ethics versus law"* – where librarians or archivists have a conflict between their ethical convictions and what they see as "unjust laws". An example of the latter was where the government instructed librarians not to buy books from a specific country.

The example of the student's reading records also hits the core aspect of *privacy*, but in a way that is typical for archives and libraries: the confidentiality of reading records, and especially *intellectual privacy* (Garoogian, 1991; Richards, 2012,2015; Rubel, 2014; van Otterlo, 2016a). The right of an individual to access and read whatever he wants *without interference or monitoring* is a fundamental requirement for intellectual growth, freedom of thought, and especially *autonomy*. When reading records are no longer confidential, people may behave differently because of social norms, oppression, or surveillance. Libraries (and archives) have always[55] protected this right, and various solutions exist (e.g. anonymity, confidentiality, privacy policies and so on) but overall it is about *principles* and how to let information flow freely in democratic societies without too much surveillance and control.

**(3.3) The Ethics of Access**

---

[55] This topic has a long history, but recently some interesting developments have happened. A group of four librarians fought a gag order by the FBI in which (under the Patriot Act) they had to hand over library records, and not even talk about it. They are called the Connecticut Four ( https://en.wikipedia.org/wiki/Connecticut_Four ) and this happened more than ten years ago. Recently they got reunited again to fight the same issues now again https://americanlibrariesmagazine.org/blogs/the-scoop/connecticut-four-librarians-fbi-overreach/

Access is the most important issue with ethical repercussions in archival practice. Preisig et al. (2014:11): *"Librarians, archivists and other information workers had to face ethical conflicts and ethical dilemmas long before digital media and the Internet started to reshape the whole information sphere. Francis Bacon's aphorism knowledge is power (scientia potentia est) refers to the fact that limited access to information and restricted education were are prerequisites of ruling elites in pre- and non-democratic societies."* Danielson (1989:53): *"Providing fair access to archives may appear to be a fundamentally simple operation, until one examines specific cases."* It often comes down to balancing many interests of stakeholders, ranging from overzealous researchers who want to gain access to legitimately privileged papers, to archivists who disagree with institutional policies, and to donors who have difficulty relinquishing control over their papers. Access problems can also come as an indirect consequence of other constraints. Danielson (1989:53): *"Due to the sheer volume of paper being generated, searches for specific documents become complicated and time consuming. The consistency of reference assistance, which can significantly reduce searching time and the attendant expense, becomes an access question."* Danielson distinguishes three distinct cases concerning access where the mechanisms of the competing demands of stakeholders can be observed.

The first concerns the *access to restricted collections*. Often there is a tension between standard archival procedures and sensitivities of donors. A simple solution is to be rather *binary* (open versus closed) in the access policy. One could be dogmatic about universal and equal access, and even promote use of records for that part of the archive that can be open. For all other parts of the archive that are sensitive or restricted, there should not be access by anyone (with a strict end date of that policy). The tension here is between freedom of information, and the privacy of the records, and people affected by the information contained. For example the famous Ludwig Witgenstein was homosexual, according to his biographer, which seemed to be withheld by the trustees of his archive containing letters. Archivists who try to retain a passive role here might get drawn into the role of mediators in the end, according to Danielson (1989, p55).

The second set of ethical issues concerns *access to open collections* and opening sensitive materials. Consequences can be simple or grow to national political proportions, for example when dealing with documents from potential war crimes. It is here that access policies can actually be indirectly influential in politics as well. Two types of access problems play a role here. One is legal: sometimes restrictions must first be lifted to obtain access. The second is "labour": to find particular knowledge in large[56] archives requires a lot of money and time, which most scholars cannot really afford. Both result, from the viewpoint of a user, in a lack of sufficient access. Furthermore, Danielson (1989, p59): *"Given, then, the undeniable tendency toward greater candor and wider access, and given the role of the archivist as arbiter in disputes over open information versus confidentiality, what are the obstacles that inhibit the development of clear standards for access policy? One obvious hindrance is backlash. ... Just as individuals are responding to a candid society with a renewed sense of privacy, so too are institutions showing a heightened awareness of security. ... Bureaucracies have become increasingly sensitive to leaks[57] of information"*.

The third set of issues concerns *equal intellectual access*, which points to particular kinds of *fairness*. In large archives it costs lots of work and money to find interesting things. One idea to help users is to inform them when researchers are after similar items. Practically it is questionable whether this works. Danielson (1989) describes several hypothetical examples related to ethics. For example, do professors get priority over the access to sources just because they are better researchers? Do fees for copy services influence the access, and should profit and non-profit making patrons pay the same fees? Should the judgment about the quality of a researcher make a difference when prioritizing access to particular still unpublished sources? And should ethical decisions be made when a journalist (who has a much faster publication medium) asks for the same information the archivist knows a researcher is working on? These are difficult issues in which the role of the archivist, as gatekeeper, as a service and as a human is important.

---

56 An interesting case here is the one on Cybersyn, the socialist big-data-avant-la-lettre project from the seventies in Chile, which was extensively described by Eden Medina in her fascinating book "Cybernetic Revolutionaries" from 2011. In 2014 Evgeny Morozov wrote a piece in the New Yorker on the exact same project (http://www.newyorker.com/magazine/2014/10/13/planning-machine). This created some controversy because some people accused Morozov of plagiarism, and quite interestingly, his rebuttal consisted of showing photographs of his own extensive search efforts in the archives of Stafford Beer (the main person in the Cybersyn project). The issue was never fully resolved (http://leevinsel.com/blog/2014/10/11/an-unresolved-issue-evgeny-morozov-the-new-yorker-and-the-perils-of-highbrow-journalism).

57 This seems similar to the increased attention to privacy issues in the library world (in the Netherlands) because of new laws about data leaks and responsibility (https://autoriteitpersoonsgegevens.nl/nl/onderwerpen/beveiliging/meldplicht-datalekken). Dutch library service organisations have just begun discussing this topic with the field (https://www.stichtingspn.nl/informatiebeveiligingprivacy).

Ferguson et al. (2016) tackles similar issues as Danielson but is based on a large empirical (interview-based) study in archives and libraries. They also focus on how professionals experience ethical dilemmas and which tools are available to assist them. Ferguson lists (among others) five dilemmas where *access* to information comes into conflict with another important *value*. The first is *censorship*. For example, archives can contain materials about groups of people which some people might see as offensive, so a balance is needed between publishing information and protecting groups. The second is *privacy*: access to information and records of that access could be in conflict if the latter need to be shared, for example with authorities. The third dilemma concerns access and *intellectual property*. The example that is mentioned here is translating something into braille without copyright compliance. The fourth conflicting value consists of *social obligations*. This one is personal for the archivist: should he or she work (partially) for free in the context of budget costs, just to maintain the level of service? The last one concerns *organisational ethos or requirements*. Here the specific case was about making university theses publicly available (with pressure for "open access") even though this might jeopardise publication of the results.

The ethics of access is about ethical issues that arise when looking at tensions between the right to information (and access to that information) and other values in the information domain. Releasing information to patrons may harm other stakeholders such as people occurring in the information in the archive, or the donor institution. Access may also be unevenly distributed, which creates unfairness and unequal access to information. Overall, it is the archivist who usually needs to deal with the ethical dilemmas: all of them involved the archivist either as a decision-maker or at least a mediator.

**(3.4) Human Professional Values and Codes of Ethics**
Given the many ethical dilemmas in accessing archives, the big question is how do archivist know how to make the right choices? Several scholars all point to the use of so-called **"code-of-ethics"** to *formalize* the values and behaviors of archivists in ethical dilemmas. A *code of ethics* formalizes rules, guidelines, canons, advisories and more for the members of a particular profession. A very early example of such a code are the *ten commandments*[58] of the Christian bible. They provide clear statements about not to worship other gods, not to kill, and not to lie. For people belonging to the religion, these are solid rules that should be obeyed, although in general most of them make a lot of sense for any individual. Another well-known example are Asimov's *three laws*[59] of *robotics,* which should be obeyed by any robot, and specify that i) a robot cannot harm a human, ii) robots need to obey the orders of the humans, and iii) a robot must protect its own existence as long as it does not conflict with the first two rules. Another influential code is the *universal declaration of human rights* which deals with fundamental ethics of human life. In previous work (van Otterlo 2014b) I analyzed this code and found several necessary alterations needed for the digital age. Recently much more interest in such issues has risen, due to advances in AI and robotics (Van Est, R. and Gerritsen, J. 2017). Yet another example, one that is more close to the archivist profession, is the code of ethics (ACM 1992) for the Association of Computing Machinery which contains general items such as *"be fair and take actions not to discriminate"* (1.4) and *"give proper credit for intellectual property"* (1.6). Such statements clearly express a desired behavior for members of the ACM. Codes of ethics can help to find ethical solutions in the sense that they formalize what is important in a profession. Codes can (Kizza 2016:27) take the form of i) *principles* (acting as guidelines), ii) *public policies* (stating norms or "acceptable behavior" in a society or a group, iii) *codes of conduct* (which may include ethical principles) and iv) *legal instruments* (which enforce good conduct through courts). Usually codes of ethics are used by high-visibility institutions and big corporations[60], but in principle any profession could define one. The main *objectives* of a

---

[58] https://en.wikipedia.org/wiki/Ten_Commandments
[59] https://en.wikipedia.org/wiki/Three_Laws_of_Robotics
[60] See for example one by IKEA (http://www.ikea.com/ms/en_JP/about_ikea/our_responsibility/iway/index.html), by Sony (https://www.sony.net/SonyInfo/csr_report/compliance/index3.html) and McDonalds

code of ethics are five-fold:
- **Disciplinary**: with discipline a group or profession can enforce professionalism and the integrity of its members. Possibly there are *penalties* for members that do no behave according to the code.
- **Advisory**: a code can help members by offering advice when difficult ethical decisions need to be made, professionally.
- **Educational**: a code can educate new members and show them the do and don'ts of the profession. Equally so, codes can educate experienced members, by polishing and refreshing their ethical knowledge in the profession.
- **Inspirational**: codes of ethics can also (indirectly) inspire members to "do the right thing".
- **Publicity**: The first four objectives are *internal usages* in which the code is used to affect (the behavior of) members of the profession. This last objective is *external* and shows the outside world that a profession has a strong code of ethics and is therefore trustworthy since its members adhere to a certain baseline of values and moral behavior.

A general distinction between different codes of ethics is whether they are *prescriptive* or *aspirational*. The former more directly prescribe the do's and don'ts, generally written as *imperatives*. Aspirational ones formalize *ideal results* that should be targeted, without necessarily telling the professional how to reach them. This resembles the difference between rule-based and consequentialist ethical reasoning. Ethical codes should not be seen as universal laws: at most they provide good guidelines but much of the practical decision-making is still up to the professional. Ferguson et al. (2016) note that codes of ethics are an important source of information for archivists, yet not always sufficient, especially not when there are *conflicts* between rules and values. Ethical codes, especially when when they have consequences when misbehaving, cause fewer discipline problems among members (Kizza 2016:50).

Archival codes of ethics have a history. The first archivists code dates from 1955, from the Society of American Archivists (SAA). It (SAA 1955) is fairly compact and states, among others, things like:
*"The Archivist should endeavour to promote access to records to the fullest extent consistent with the public interest, but he should carefully observe any proper restrictions on the use of records".*
Similar statement come from the universal declaration on archives (ICA-DL 2011):
*"Archives are made accessible to everyone, while respecting the pertinent laws and the rights of individuals, creators, owners and users".*
*"The Archivist should respond courteously and with a spirit of helpfulness to reference requests."*
*"The Archivist should not profit from any commercial exploitation of the records in his custody."*
The first two deal with the same ethical issues I discussed in the previous sections, but the third resonates a worry that was expressed before that. Torres-Vargas (2005, 158) *"One of the risks identified by Wells was the publication of the World Encyclopedia by a commercial publisher, with overriding financial interests."*
The 1992 version (SAA 1992) extended the previous code (with an intermediate version in 1980) and added *commentaries* to explain the parts. It also added a text explaining the objectives of the code, such as the ones discussed above. The text on access now includes:
*"It is not sufficient for archivists to hold and preserve materials: they also facilitate the use of their collections and make them known."*
This amounts to the preservation, use and publicity aspects of the archive. It also contains:
*"Archivists endeavour to inform users of parallel research by others using the same materials, and, if the individuals concerned agree, supply each name to the other party."*
This refers to a dilemma I have discussed earlier in this section.
The final commentary of the code states something about potential conflicts:
*"When there are apparent conflicts between such goals and either the policies of some institutions or the practices of some archivists, all interested parties should refer to this code of ethics and the judgment of experienced archivists."*
In a subsequent version (SAA 2005) all commentary was removed, apparently because it was deemed less useful and the SAA might be subject to legal liability. The most recent version however, (SAA 2012) is conceptually very different. In this version there are two main components. One is the code of ethics as before, but the other is an explicit statement of the *core values* of the archival profession. According to the introduction paragraph, the latter represents what the archivists *believe* while the former represents a framework for the archivists' *behavior*. This division in values in behavior is intuitive and could be a way to

(http://corporate.mcdonalds.com/mcd/investors/corporate-governance/codes-of-conduct.html)

solve some of the ethical dilemmas I discussed before. The values drive the behaviors, but if values are stated separately, they may be useful input for a utilitarian analysis of a particular ethical dilemma. Core values in the document are about access and use, accountability, preservation and more. For access it emphasizes that access to records is essential in personal, academic, business and government settings, and use of records should be welcomed. Later in the code of ethics itself this value is translated[61] into *"minimize restrictions and maximize ease of access"*.

A Dutch version was issued by the Dutch royal association for archivists (KVAN 1997) and although it maintains a different structure, it captures the same kind of values. Related codes exists for libraries, such as the professional charter for librarians in public libraries (PL 1993), and codes by the American library organization (ALA 2008) and the International Federation of Library Associations and Institutions (IFLA 2012). As explained before, libraries do have different activities, but the core values are shared with archivists, which can be seen in the similarities with library values concerning access. Libraries often explicitly state that they *promote information literacy* though (IFLA 2012). In addition, libraries often want to state that *"if certain information is against the personal view or conviction of the librarian, this will not influence the evaluation of the information."*, making explicit that they do not judge upon information requests – something that relates to the mystery shopper case I described in the introduction. In addition, like I have described earlier, archives, libraries and museums are similar if we look at the preservation and provision of cultural information. The ICOM (2013) code of ethics for museums therefore is very similar, and highly related, to archival codes. Instead of values, the code first states *principles* after which a detailed set of guidelines is given, ranging from access to acquisition policies. Occasionally separate codes are made with respect to specific aspects such as privacy[62], for example as was done recently by IFLA in 2015.

One main issue with codes of conduct is that guidelines can be *non-committal*. The difference between prescriptive and aspirational version is large, but still the real dilemmas are caused when multiple values or rules conflict in a practical setting where the archivist needs to make a decision. Several authors do see the benefit of codes of ethics (Preisig et al. 2014; Cox 2008; Ferguson et al. 2014; Danielson 1989) but they also stress these shortcomings, and call for balance between various influences. As another solution, Morris[63] calls for an *enforceable* code of ethics, just like legal and medical professions are governed by codes of ethics which carry the force of the law. Violations are subject to *sanctions* including loss of license and civil and criminal liabilities. Enforceable codes are desirable because they would raise professionalism, and be way to earn the trust of the public to maintain the role of "guardians of the archive". One issue that needs to be resolved still is whether the SAA would be the regulating body or someone else. Overall, enforceable codes would require more consensus on the exact set of ethical values and rules, thereby removing some of the decision freedom of the individual archivist.

Formalizing ethical codes, core values and guidelines has one main purpose: to formalize *how humans should behave*, in this case in the archival profession. This is all about *human values* and *human behavior*. Ethical codes appeal to human values, human emotions, and human practical decision making. By formalizing it in a code it has become *transparent* and can be communicated to peers, users, donor organisations and the general public.

### (4) Transformation to the Digital World
The physical world I have described in the previous sections, together with all ethical norms and systems that have evolved, are in transition. The digital era is here and slowly *digitalizes* all that was once physical. In addition, *algorithms* will intelligently, and omnipotently govern all interactions that were once physical. The new norms that are needed in this new world are the topic of the next section. Here I will first describe the transformation by algorithms in society at large, and in archival and library contexts in particular.

### (4.1) Digitalization and Algorithmization
Despite digital technology being around for decades, **digitalization** is currently a hot topic in business and technology. The transformation of products, manufacturing, marketing and customer relations through digital technologies comes with huge expectations for new commercial opportunities. One of the biggest hype terms for the current decennium[64] seems to be *big data*. Everywhere around us everything is turned into *data* which is thought to be good for health, for the economy, for personal well-being, for the advancement of knowledge, and so on (Viktor Mayer-Schoenberger 2013). The main promise is that if we gather and use lots of data about anything, we will understand it better, make better predictions about a domain, and be better in *optimizing* our policies and strategies. For example, by gathering lots of patient data, and by building *statistical models* to predict diseases, and by *experimenting* with novel treatments based on the insights of data, we will be able to cure more diseases. A major point made by many people is that the age of big data allows one to throw[65] away typical "hypothesis-driven" science, which works *top-down*, and to adopt a more *bottom-up* strategy, which starts with the data and tries to find patterns in there and then generalizes.

---
61 Which sounds very much like a *Utilitarian* maximization.
62 https://www.ifla.org/node/9803
63 http://slanynews.blogspot.nl/2010/09/enforceable-code-of-ethics-why.html

For many people the age of big data seems rather new, but that is not entirely true (and see also (boyd and Crawford, 2011) for some general critique on big data). Technical fields such as computer science and mathematics have always been doing big data research, but the main new things are developments (van Otterlo and Feldberg, 2016) in hardware and (open) software, and the growth of data in general, that make large-scale data processing possible, for anyone. Big data "avant-la-lettre" can for example be found in the *Cybersyn project* in Chile in the seventies which was aimed at controlling the economy of a complete country, something which sounds like modern "smart city"[66] endeavours. Cybersyn was based on *cybernetics* (Pickering 2010), a precursor to modern AI. Medina (2015) brilliantly contrasts Cybersyn with modern big data enthusiasm and describes five lessons to be learnt from this old project, emphasizing the *sociotechnical* aspects, contrasting that with *technological determinism*. Data has always[67] been gathered since it was technically possible, but the scale of today is indeed huge.

Modern data-driven directions can be seen as a new[68] machine age, an industrial revolution (see also Floridi, 2014). After the rationalization of both human labour and cognitive labour, we now enter a new phase where much of our society gets turned into data, and processed by autonomous, artificial entities that will take over a lot of activities from humans. Predictions of where this will end are plenty, and a fairly cautious one is given by Butler (2016) who identifies *drivers,* and *enablers,* such as increasing computing power and computation speed.

The *datafication* of our world is a huge development, and slowly scientific literature is created in which its consequences are studied (see (van Otterlo 2013,2014a,2014b,2016a,2016b) and cited literature). For this text, it is important to look at *how* we get from a purely-physical world to an increasingly more digital version, and what the consequences are for our society in general, but more specifically for the physical archives and archivists I have introduced in the previous sections.

In the figure, each blue square in the blue area represents an *object,* each green triangle a *document* and each red circle a *person*. Traditionally, all relations and interactions between any of these groups was *physical* as I have described before. In our modern age, all such interactions are becoming *digitalized* step-by-step and produce data entering the red area. If we consider shopping, long ago, one could go to a store, fit some jeans, pay them and no one but the sales person (and the customer) would have a faint memory of who just bought which jeans. Nowadays, traces of security cameras, online search behavior on the store's website, WiFi-tracking in the store, and the final payment, all generate a *data trace* of all interactions with the store and its products. A major consequence of that digitalization process is that a *permanent memory* of all those specific interactions is stored in a *cloud* and can never be forgotten. In addition often this data is generated and governed by *private entities*. For example, Facebook governs a lot of our social interactions on the their platform and keeps data about us, Google gathers everything that people do with its search engine, and Twitter keeps score of all our quick interactions via Tweets. This data is to some extent owned by these companies (see also Waller (2009)), and whereas a long time ago interactions were physical, and no trace was kept, these modern platforms are *aimed* at gathering as much data as possible of all our interactions, and aimed at *retrieval* of that data (of all users combined) for purposes such as *profit* and *surveillance*. Big companies govern not only our information behavior through social networks and search engines, but they also govern our entertainment activities (e.g. Netflix), our love life (e.g. Tinder, OkCupid), our museum visits (see Meermanno[69] for an example), our interpersonal communication (e.g. Gmail, Twitter, WhatsApp), and more.

---

64 The start of this direction was only roughly ten years ago
The Petabyte Age https://www.wired.com/2008/06/pb-intro/ (Mitchell 2009)
Mining Our Reality http://www.cs.cmu.edu/~tom/pubs/Science2009_perspective.pdf (Anderson 2008)
65 This phenomenon is called "the end of theory" since it breaks with standard scientific methodology.
66 See for example Barcelona (http://www.smartcityexpo.com/barcelona) and other cities.
67 See for example East-Germany's Stasi and the great movie about it http://www.imdb.com/title/tt0405094/
68 See the Rathenau Report on "Working in the Robot Society (2015) https://www.rathenau.nl/nl/node/766
The Rathenau Insitute publishes many reports on the digital society and its implications, see
https://www.rathenau.nl/nl/publicaties

A good example of this transformation from the physical world to the digital comes from *photography*. Vivian Maier[70] was a fantastic street photographer of the old, non-digital world. Her film-work was discovered by John Maloof, who bought some of her stuff in a sale and found many undeveloped film roles. She herself had not seen thousands of her own photos, and they existed only as unique physical objects which were almost thrown away, had it not been for Maloof. Contemporary photography is so much different. Kai Man Wong[71] is a photography expert and vlogger who makes movies for Youtube. Everything he does with a camera, or about cameras, is digital. His photos, his movies, his Twitter account in which he announces new videos, his editing tools on his laptop for movies and photos: all digital. Even without him actively pursuing it, a giant *archive* is being created, in real-time, as a *memory* of all this (inter)actions, by all these big corporations that make Youtube, cloud storage, Twitter, online video editing software, and so on. Digitalization of Maiers work has now inserted her in our collective digital memory, but how many undiscovered photographers of the physical world are truly forgotten? Once physical objects are destroyed, they seize to exist, unless they are drawn into our digital memory.

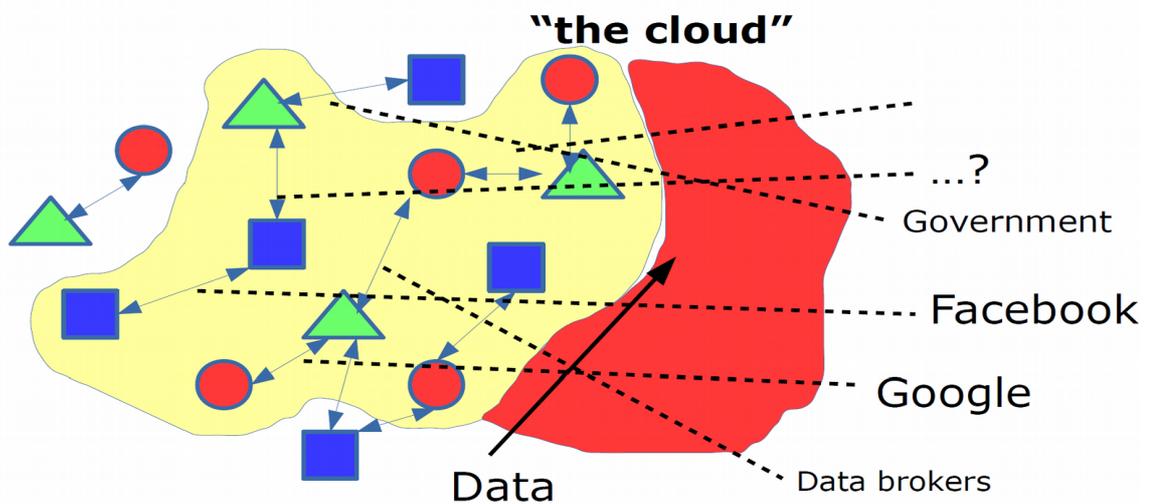

Another good example concerns *reading*. So-called *e-readers* have made it much easier to bring virtually unlimited numbers of books on your holiday trip, in contrast to a long time ago when only a handful of physical books would fit your luggage. However, new technologies require serious rethinking of codes which were devised in an age dominated by print (Ferguson et al. 2016). Digital books have many more consequences than just the direct ones of carrying capacity. The digitalization of reading activities creates digital traces of reading, browsing, times, length and places of reading activities, markups, sharing, book finishing and much more. Such traces can for example be used in new business models for the publication of books and royalty payments to authors, all based on exact data how "well" a book is read or received by the general public. One can for example pay an author based on how many pages are actually read[72]. This may raise questions about *fairness* in digital book prices. Besides that, already the "having" of a book changes with digitalization. For example, a PDF file containing the text of a book could be considered a local copy that the user "owns" as a property, but more in general, digital copies of books on e-readers should be viewed as a *"right to access"*[73] in the same way people have access to entertainment on NetFlix and Spotify. Because of this, digital books are again more like *physical* libraries or archives: the user gets (temporary) *access* to an item based on prior payment or an ongoing subscription. An explicit demonstration of this was when Amazon (accidentally?) revoked access to several books – ironically Orwell's 1984[74] was one of them – which affected many readers, *while* reading the book. Digitalization here changes *property* into *access*, with extra consequences being that all aspects of that access will be monitored. Baron (2015) describes two *tests* to distinguish between the physical and beyond (the digital). One is the *ownership test* which asks for whether one can actually own something, like a book, and put a mark on it (for example, an *ex libris* stamp). The other one is the *autograph test* which is similar and asks for whether one can bring a copy of a book to a signing session to obtain an autograph (for example of the author) on the physical copy. It goes without saying that modern e-books fail both tests; Baron (2015:133) quotes Michael Dirda: *"E-books resemble motel rooms – bland and efficient. Books are home – real physical things you can love and cherish."*

---

69 The Dutch book museum Meermanno (https://www.meermanno.nl/) recently featured the *Conn3ct* exposition (http://conn3ct.media/), in which the physical was contrasted with the digital. Visitors were given an RFID card with which they could interact with the collection in various ways. Information contributed by visitors was shared with other visitors through various screens, displaying information like *"You chose answer X to this question… 68 percent of all visitors today chose that answer"*, thereby engaging the public in new, digital ways by utilizing the physical aspects of the museum.
70 https://en.wikipedia.org/wiki/Vivian_Maier
71 https://www.youtube.com/channel/UCknMR7NOY6ZKcVbyzOxQPhw

Another aspect of the digitalization of reading that goes beyond monitoring and tracking is *social reading*, which relates to all (digital) ways to share reading activities *with other people*. Winget (2013) describes four categories: talking to a friend about a book, discussing a book online, formal book clubs, and engaging in discussions in the margin. According to Winget, scholarly work on social reading is limited so far, yet practically many systems now support social reading. For example sites like GoodReads, and hardware like Amazon's Kindle and the Kobo e-reader, allow users to indirectly "communicate" by making notes in books which can get shared. For example Kindle features *popular highlights* which are passages in a text that got highlighted by at least a couple of people. Such *frictionless sharing* (Richards 2012) brings us new ethical challenges. For example, is my reading experience different if I know that *"80 percent of all readers does not finish the book"*? Do readers get a biased view when they receive the popular highlights before forming their own opinion? Are people aware that their markups of highlights can be shared instantaneously with many other readers? All these aspects severely the impact (intellectual) privacy of the reader.

*Privacy* is a pressing item in (social) reading (Lynch 2017). In the physical library, privacy is well-defined in the same intuitive terms most people talk about privacy. Sommer (1969, p45): *"Few places make as strict a demand upon the physical setting to guarantee privacy as the library reading area. It is one of the few places where interaction between people is actively discouraged."* However, the complete opposite is now starting to become the norm because of all the sharing many internet platform encourage, including social reading platforms, including Twitter and FaceBook where people can broadcast what they are reading and what they are thinking of it. Not long ago, reading records were (in essence like any thing people now share openly on FaceBook) very confidential[75] and librarians were very aware of that. The Patriot act and related regulation have now made those records more vulnerable[76], but with developments such as social reading it seems like a lost case. For example, when buying at Amazon you give away what you buy, but with e-reading on the Kindle there seems not to be any way to read without being tracked, or Baron (2015, p150) *"with e-books all privacy bets are off"*. As Alter (2012) writes: *"Your e-book is reading you."*

---

72 There are many opportunities and incentives for book publishers and sellers to create systems in which quantified aspects of "reading" are coupled to (monetary reward for) "writing". Amazon's new way of paying can be based on the data obtained from their Kindle e-reader (http://fortune.com/2015/06/24/amazon-kindle-authors/).
73 This falls under the *digital rights management* (DRM) technologies (https://en.wikipedia.org/wiki/Digital_rights_management)
74 https://www.theguardian.com/technology/2009/jul/17/amazon-kindle-1984
75 https://www.theguardian.com/world/2015/jun/05/nsa-surveillance-librarians-privacy
http://www.ala.org/advocacy/advleg/federallegislation/theusapatriotact
76 http://www.npr.org/news/specials/patriotact/patriotactdeal.html

In recent analysis on privacy and social reading Jones and Janes (2012) take Nissenbaum's *contextual integrity* framework and derive recommendations in the context of legal frameworks. Richards (2012, 2015) on the other hand, points to the perils of social reading and sees the intuitive hazard for what he calls *intellectual privacy*: the idea that for some ideas people need to know that they are not being watched. It resembles the typical Panopticon[77] style of privacy concern where people change their behavior if only they know that they might be watched. Basically, frictionless sharing, i.e. where all information is automatically shared and disclosed, is actually completely *not* frictionless from a privacy point of view. Note that not all sharing is bad; in fact, discussions about books can be beneficial to all parties. The problem is that one needs to be able to select and read in complete privacy if wanted, and that becomes harder each day. Lynch's (2017) recent study of reader privacy contains many of these elements and features case studies in *student-textbook interaction tracking* and *government interests in reading habits*.

In contrast to what many people think, *data* is not the most important entity in our digital world. Data is only a *consumable* for the entities that really change our world: *algorithms*. Algorithms are *computer programs* that autonomously utilize data in order to *do* something. This can be *sorting* names in a database, computing navigation instructions, but also organizing Facebook's news feed, analyzing people's e-book reading habits, and recognizing faces from an Instagram photo feed. The term algorithm[78] stands for any *finite procedure/recipe*, with *well-defined instructions* and which is *effective* in solving a problem. Algorithms can exist outside the digital world: a detailed recipe can be considered an algorithm too, and so can instructions for setting a breakfast table, provided that the individual steps are clear enough. In this text I focus on digital algorithms, which increasingly become more *complex*, *intelligent* and *adaptive*. The **algorithmization** is the process of carrying out increasing numbers of tasks in society using algorithms. The main field studying and creating such algorithms is a subfield of computer science: AI.

AI[79] (McCarthy, 2007; Russell and Norvig, 2009; Poole and Mackworth, 2010; Nilsson, 2010) as a field is more than 60 years young and has a past in cybernetics (Pickering 2010). AI has been developing intelligent algorithms for a long time, but in recent years its progression has exploded, under the influence of an enormous growth of the *tech* industry, a wide-spread availability of data and computing power and several breakthrough technologies such as reinforcement learning (Wiering and van Otterlo 2012) and *deep learning*. Some earlier success of AI may have involved *deception* in order to create an *illusion* of intelligence *(*Sharkey and Sharkey, 2006) but lately AI has shown impressive progress on a number of domains such as computer vision, natural language understanding, and robotics. The core of AI is coming up with intelligent systems that *in some way* exhibit observable behavior for which some form of intelligence is required. Traditionally this was about *reasoning, planning, and mathematical problem solving*, but in the current data age, it is about *learning.* The subfield of AI called *machine learning* (ML) (Flach, 2012; Domingos 2012, 2015) specifically focuses on using data to learn how a particular task should be performed, which can range from baking cookies[80] to driving autonomous cars by learning from popular computer games[81]. AI is rapidly becoming *the* driver for innovation[82].

---

77 https://www.theguardian.com/technology/2015/jul/23/panopticon-digital-surveillance-jeremy-bentham
78 https://en.wikipedia.org/wiki/Algorithm
79 *Science*, special issue on how A.I. is transforming science
http://science.sciencemag.org/content/357/6346/
80 http://www.wired.co.uk/article/google-vizier-black-box-optimisation-machine-learning-cookies
81 https://www.technologyreview.com/s/602317/self-driving-cars-can-learn-a-lot-by-playing-grand-theft-auto/
82 Goverments, universities and large companies are assessing the sudden rise of the powerful technology of A.I. See for example the report from the field itself (Stone et al, 2016) and some other examples such as:
Andrew C. Scott José R. Solórzano Jonathan D. Moyer Barry B. Hughes (2017) Modeling Artificial Intelligence and Exploring its Impact, working paper, Pardee Center for International Futures
http://pardee.du.edu/sites/default/files/ArtificialIntelligenceIntegratedPaper_V6_clean.pdf
Executive Office of the President National Science and Technology Council Committee on Technology (2016) Preparing for the Future of Artificial Intelligence (2016)
https://obamawhitehouse.archives.gov/sites/default/files/whitehouse_files/microsites/ostp/NSTC/preparing_for_the_future_of_ai.pdf
Jacques Bughin, Eric Hazan, Sree Ramaswamy, Michael Chui, Tera Allas, Peter Dahlström, Nicolaus Henke, Monica Trench (2017) Artificial Intelligence: The Next Digital Frontier, discussion paper, McKinsey Global Institute.

The most promising development in recent years are the accomplishments of the British company Deepmind (now part of Google/Alphabet). Metz (2016) describes how an adaptive, intelligent computer program called

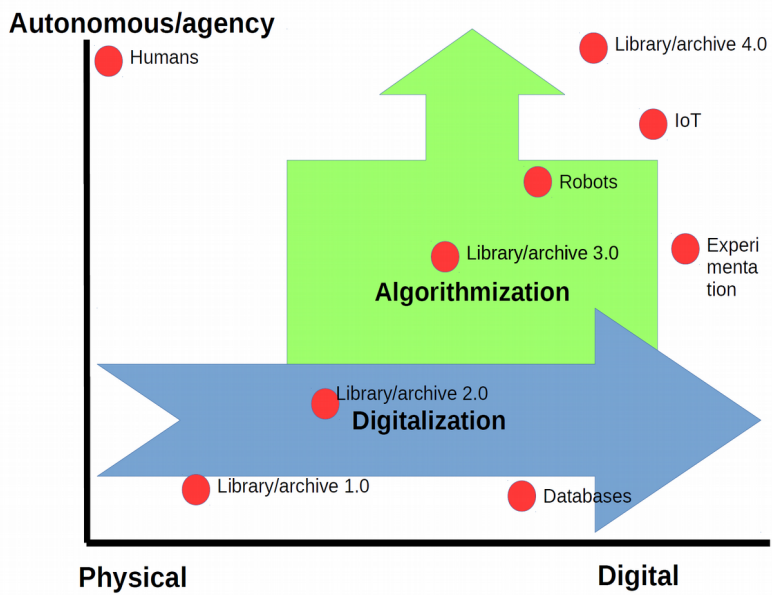

*AlphaGo* managed to beat the human champion in the game of *Go* (a board game with a much higher complexity than chess, of which the human champion Gary Kasparov was already beaten in 1997 by IBM's DeepBlue computer). The program learns from existing games, and adaptively trains itself to become better. The general structure of the approach lends itself to many other situations, for example to control popular, computer games such as Starcraft-2[83], but it could also manage a city, or maybe in the end, the world itself.

The transformation into a digital society can thus be characterized by the two interrelating developments I have discussed in the previous two subsections: *digitalization,* which turns once-physical interactions into digital *data,* and *algorithmization*, which amounts to increasing analysis and utilization of that data by algorithms. Together they characterize two dimensions with which the **transformation to a digitalized world** can be characterized as also shown in the figure. The first dimension, *digitalization*, represents the process of making physical interactions digital and is depicted on the X-axis. The more to the right, the more aspects of society have been digitalized. For example, traditional libraries and archives are positioned far to the left, whereas our banking and financial systems has become increasingly digital and can be found further to the right. Relatively novel developments in digitalization are *the internet of things* (IOT) (Ng and Wakenshaw, 2017) and *robots.* IOT makes ordinary things such as speakers and refrigerators "smart" and digital by letting them communicate with other devices, and creates *sensor-based* environments[84] where physical actions of humans can be tracked using cameras and other types of *sensors.* IOT represents an advanced step which digitalizes even more aspects of the physical world. Robots on the other hand, represent a completely new step in which new smart, digital "entities" are introduced in the physical world that interact in human-like or other ways. The *physical presence* of robots raises many new issues compared to digital *databases.*

The second dimension, *algorithmization*, concerns employing algorithms to utilize data for various purposes. Note that algorithmization is partially dependent on digitalization: more data means more opportunities to analyse it. However, algorithmization also has its own dynamics; advances in the algorithmic state-of-the-art can create entirely novel technologies that in themselves may also cause more data to be produced. For example face recognition is[85] now widely used on social networks such as Facebook, which results in more (meta-)data about photos and with that more detailed knowledge about social relations. Currently so-called *deep learning* is an *enabler* for many novel applications such as image recognition and various *prediction* tasks but possibly in the near future *quantum computing* will cause an even bigger revolution in terms of algorithmic *capabilities*. The Y-axis order can be qualified in terms of "how advanced" the algorithms

---

http://www.mckinsey.com/~/media/McKinsey/Industries/Advanced%20Electronics/Our%20Insights/How%20artificial%20intelligence%20can%20deliver%20real%20value%20to%20companies/MGI-Artificial-Intelligence-Discussion-paper.ashx

83 https://deepmind.com/blog/deepmind-and-blizzard-release-starcraft-ii-ai-research-environment/

84 See for example BLIIPS project which deals with a sensor-based, brick-and-mortar *public library* (van Otterlo, 2016b/2017; van Otterlo and Warnaar 2017) and also other *instore tracking* technologies such as the Amazon Go supermarket where all purchases are detected using sensors and automatically deducted from one's credit card (see further https://www.amazon.com/b?node=16008589011).

85  https://www.theverge.com/2017/3/15/14937982/facezam-facial-recognition-facebook-photos-privacy

employed are. For example, a simple Excel-based computation of customer data characteristics will end up lower on the Y-axis than a care robot running around in a hospital. However, a more precise ordering of algorithms is to look at their *autonomy,* or *level of agency.* Compared to a simple Excel-sheet executed by a human operator, the care robot or a fully automated trading algorithm has more agency: they gather data, analyze it and, most importantly, *act* upon that data without much interference of human operators.

**(4.2) Digitalized Archives and Libraries**
The influence of digitalization and algorithmization in the archival and library domain are potentially huge. Equally so, our current digital age contains many "rival" information providing services on the internet. *"Can libraries actually deliver different vehicles for serendipitous discovery if everyone is getting from A to B using a search engine?"*[86] But, as Zastrow notes (2013:17): *"Although rarely discussed in the professional literature anymore, easy access to full-text online sources has created a kind of McLibrary"*. The International Federation of Library Associations (IFLA) recently published a trend report on five major transformations in society (that could affect libraries). Trend 1 is that *"new technologies will both expand and limit who has access to information"*, Trend 3 that *"the boundaries of privacy and data protection will be redefined"* and Trend 5 says that *"the global information economy will be transformed by new technologies"*. Archives and libraries in the digital world face changes because of a changing role in the landscape with new providers of (digital) information such as Google and Wikipedia, but also because their collections are being *digitalized* and their core operations will be *algorithmized*.

Much has been written about the "future of the library" or archive for that matter, for example by Licklider (1965) and Lancaster (1982). Logan and McLuhan (2016) identify principled *laws of the medium* which describe how the medium, for example electricity or digital, has an impact on how aspects are changed, gained or lost when that medium is changed. Other recent texts have argued for why libraries are still highly relevant in our digital age (Herring, 2014; Palfrey, 2015) while others have defended the physical books[87] In general, libraries and archives (to some extent) have always struggled with their exact role, especially in the transformation to our digital age, with the novel aspects of born-digital records and books, and with the loss of being an information providing monopolist (see also Anderson 2011:212) now that Google and co have appeared. Both Kallberg (2012) and Clement (2013) have investigated how the archival profession changes in our digital age, and how archivists think about that transformation. Paulus (2011) shows that the *lifecycle of information* of archives and libraries changes, and that, for example, a transformation is happening in which libraries return to an ancient and medieval model of the library as a site of both production and preservation. Cox (2007): *"At last, archives have a real opportunity to abandon the role of gatekeeper and invite user participation, interaction, and knowledge–sharing."*. He continues " *What would happen if we could engage our users in defining and describing archival content and in communicating it to others? Is it possible that the analog archives tradition can learn from the movement of social media and social design? Some of the opportunities include diminishing the role of the archivist as gatekeeper, promoting participation and collaboration among users, and enriching the archives itself by tapping into the specialized and diverse knowledge of researchers"*. Overall, digitalization (and algorithmization as a second step) has an impact on the meaning, role in society, and functioning, of libraries and archives, and with that, the role of the librarian and archivist, see for example the *atlas for new librarianship* (Lankes 2011).

The future of the library has many parallels with the development of information technology such as the internet. Terms such as "web 2.0" or "web 3.0" can be mapped to developments in libraries, and archives. Both Noh (2015) and Kwanya et al. (2012) describe several stages leading up to "library 4.0", which is where *"technology will become one with users' lives"* (Noh 2015). Currently we are in the "3.0." phase in line with "web 3.0." where data and analysis is performed through AI, and where all knowledge becomes connected by the *semantic web*. Before, stage "2.0." was defined by the "social web" (for humans, whereas the semantic web is for machines) and stage "1.0" for desktop PCs and the World Wide Web (WWW). The main conditions for constructing the "web 4.0.", and with that libraries and archives in that stage, are *ubiquity* (digital and offline become blurred), *identity* (protocols know who you are) and *connection* (everything is continuously connected). Noh (2015): *"However, library 4.0. must not only include software-based approaches, but also other technology such as makerspaces, Google Glass, context-aware technology, digitalization of contents, big data, cloud computing and augmented reality."*

---

86 The International Federation of Library Associations (IFLA) published a trend report in 2013, with an update in 2016, on on several big developments, see https://trends.ifla.org/
87 See for example the interesting debate between Umberto Eco and Jean-Claude Carriere in *"N'esperez pas vous debarrasser de livres"* *("This is not the end of the book")* from 2009.

Other developments can be found in the core operations: 1) selecting (and discarding) which items belong in the archive, 2) ordering and physically arranging the records in a physical layout, and 3) describing and documenting which records exist and where they are. Recently, scholars have looked at how digital technologies (both data and algorithms) can have an impact on these core operations and beyond. Both Fernandez (2016) and van Otterlo (2016b, 2017) describe how AI can be employed to do, for example, book recommendations based on access to items and user data: (Fernandez, 2016:21): *"So, when Amazon's AI makes a recommendation that you might like a particular author based on the last book you purchased, it is drawn not only knowledge about you, but also from information minder from the millions of other users."* AI can be used for many things, ranging from retrieval of sources to personal assistants implementing virtual reference desks, and to *optimizing* various library and archival processes. In principle, all of the core archival processes can be *automated* but currently digitalization has only gotten so far, and algorithmization still has to start really. And again, since archives are more idiosyncratic (than libraries) in terms of how they select, order and catalogue, chances are that it will take longer before algorithmization takes over here. Zastrow (2013:18): *"The idiosyncratic and contextualized world of archives necessitates communication with the archivist. Because archival collections are not composed of discrete, individually cataloged items, disintermediation does a disservice to our scholars."*

When it comes to digitalization, libraries and archives are in transformation. Collections are constantly being digitalized to provide more wider public access to information, for example through the American project *Digital Public Library of America* (DPLA[88]) and the European counterpart *Europeana*[89]. These digital initiatives unlock massive amounts of archival data such as books, photographs and various documents. Initiatives such as the Google Books project are similar in terms of technology, but have different goals. Google Books[90] has a long history of battles[91,92,93] between a tech giant wanting to unlock *all* books written by mankind, for everyone, and author organizations that think that Google does not have the *right* to do that in this way. The ethical issues of access here are severe, since Google may want to push the idea being a *universal librarian* but many think this role should not be pursued by a commercial entity whose business model is based on advertisements, algorithmic rankings, and competition between information sources. Instead, many argue, such projects should be governed by a *public institution*, which is one of the reasons projects such as Europeana and DPLA were founded. In general, making collections and catalogues digital is something that is happening for decades already, long before the *big data* age. Also, public libraries have been logging user transactions for a long time and have created large databases that can be used for collection analysis and optimization, but also increasingly so for targeted marketing[94] purposes and more (see Allison 2013). Novel developments in the Netherlands are initiatives like a *national collection plan*[95] and the ongoing construction of a *nationwide* database (NBC+[96]) of all *transactions, user records and book*

---

88 http://dp.la
89 http://www.europeana.eu
90 http://books.google.com
91 https://www.wired.com/2017/04/how-google-book-search-got-lost/
92 https://www.theatlantic.com/technology/archive/2017/04/the-tragedy-of-google-books/523320/
93 See https://www.theatlantic.com/technology/archive/2015/10/fair-use-transformative-leval-google-books/411058/ and https://www.wired.com/2017/04/how-google-book-search-got-lost/
94 See the (Dutch) report on customer groups and profiles, and how to approach different groups in the customer base of public libraries (http://www.debibliotheken.nl/fileadmin/documenten/vereniging/2008_de-klant-is-koningin_handboek_openbare_bibliotheken.pdf). Nowadays, public libraries themselves are utilizing their data assets to improve various library services.
95 See the (Dutch) report "Gezamenlijk collectieplan Beleidskader collectiebeleid voor het netwerk van openbare bibliotheekvoorzieningen" (December 2016) on the formation of a nation-wide collection for public libraries. (https://www.kb.nl/sites/default/files/docs/191216_gez._collectieplan_definitief.pdf)
96 See the NBC+ (National public library catalogue) which contains the full collection of all Dutch public libraries, and in augmented with a data warehouse storing transaction and patron records for a large percentage of libraries. This creates one big database with core information about the public library in the *information* in public libraries. So far, only public access to the collection information has been arranged. However, the full database will eventually be useful for algorithmic innovations such as *recommender systems* computing personalized book suggestions for individual patrons.

A novel way of digitalization is to digitalize interactions that were purely physical until very recently. Cox (2007) makes the case for "machines in the archive", such as photocopiers[97] and (personal) cameras with which users can digitalize materials themselves. The BLIIPS[98] project (van Otterlo 2016b) represents an effort to make the *public* library more intelligent through AI. One of the goals is to use *sensors* to digitalize physical actions such as moving around through the building and interacting with the physical collection (e.g. scanning a book cover to look up information in the catalog). The approach makes use of users' smartphones, and makes some library services such as looking up a book *context-aware*: based on the current location of the user, the smartphone can provide real-time navigation, similar to Google Maps, but now *indoors* (van Otterlo and Warnaar 2017). The overall goal is to make the physical library behave like Google: you type in a search query and the system will lead you to the book. BLIIPS is based on a combination of ideas coming from digitalization, AI, retail strategies (e.g. customer journeys) and experimentation based on data science, and is aimed at creating a physical library laboratory in which algorithms can optimize library services based on data. Several similar approaches target the interplay between physical and digital worlds in libraries, such as in the *fused library* (Buchanan 2010), *blurred boundaries* (Walsh 2011) and *location-based recommendations* (Hahn, 2011). Other efforts in libraries have been focused mainly on *conversational agents* which can act as guide or tutor to assist users in query formulation, or as a general *avatar* representing desk personnel (Liu 2011; Talley 2016).

Digitalization will go hand in hand with centralization (gathering more data sources and combining them) and opening up on the internet (providing access for many people). Ethical challenges already start here, since deciding whether to digitalize anything in the first place (and discarding a possibly unique original) amounts to an ethically loaden decision. The Google Books project exemplified the ethical problems with privacy and copyright when digitalizing and unlocking and also projects like BLIIPS show that very easily ethical challenges related to monitoring, surveillance (Hellman, 2015) and experimentation pop up when digitalizing people's private, physical interactions. In the next section I will address general ethical issues when algorithms come into play.

### (5) The Ethics of Algorithms
In this section I turn to the ethical implications of algorithmization. This can include old ethical dilemmas in a new, digital world, or new ones raised by algorithms. This section features two different types of *taxonomies* to identify ethical issues with algorithms.

### (5.1) Algorithms are Biased Codes
As I discussed previously, algorithmization takes over many tasks in society that were once physical. This creates opportunities for *technological innovation*, ranging from simple database operations to robotic surgery. People often associate with algorithms properties such as *infallible*, *trustworthy*, *exact*, and especially: *objective*. Because computer-based algorithms are based on *mathematics* (logic and statistics) people tend to think that *because of that* algorithms are objective and fair, since they can compute *the best answers* given the data. While some of this may be true, in general algorithms are far from objective: they are heavily *biased* (Bozdag 2013; van Otterlo, 2013) Consider for example (part of) a simple algorithm for a bank, specifying that *"IF sex = female AND age > 60 THEN decision = no-life-insurance-policy"*. Now this algorithm is perfectly mathematical, and exact, and it thoroughly computes from personal data whether somebody is eligible for a life insurance policy. However, from a human point of view, it is far from "objective", or "fair" since it discriminates against women above 60 years old. Its decisions are *biased* and it discriminates, in plain sight. To make things worse, we can also imagine a second algorithm which is specified as *"IF f(sex) * g(age) > 3.78 THEN decision = no-life-insurance-policy"*, and let us assume it makes exactly the same decisions as the first. A problem here is that this algorithm discriminates too, but it is

---

Netherlands (https://www.kb.nl/ob/digitale-infrastructuur/nationale-bibliotheekcatalogus-nbc).
97  This also connects back to Eco's "restrictions" described earlier on being able to photocopy in a hostile library, but also to the ethical challenges concerning fairness when photocopying costs money.
98  http://martijnvanotterlo.nl/bliips.html

hard to see from its description because we do not know what the functions f() and g() do, and also not why there is a threshold of exactly 3.78. Maybe these aspects have been *learned from data* which would require us to have a look at the data and learning process to form an opinion about the algorithm's objectiveness.

To analyze bias, one can take *formal* notions and look at how bias can be computed[99] from an algorithm, or how it affects *adaptive processes* such as ML (and the *bias-variance trade-off*, see Flach 2012, van Otterlo 2013), or how bias affects many parts of information providing services (Bozdag 2013). A simpler illustration of the role of bias, and how it relates to choices or *intentions* of the creators, was given by Birkbak and Carlsen (2016a). They take as a starting point the main intentions of algorithms as communicated by Google (*"organize the world's information and make it universally accessible and useful"*), Facebook (*"to give people the power to share and make the world more open and connected"*) and Twitter (*"create a global conversation"*). They operationalize these by constructing algorithms that use these intentions on a retrieval task: to rank a small set of scientific documents. First, Google's operationalization of the ranking algorithm prioritizes the amount of citations, because this mimics Google's PageRank algorithm that considers items more popular if more sources refer to it. The Facebook version prioritizes citations from co-authors, since its algorithm makes likes (citations) from friends (co-authors) more important. Finally, the Twitter version prioritizes citations (retweets) from non-friends (non co-authors). They then show how such built-in (ethical) values (bias) create *different* rankings of the same documents. Birkbak and Carlsen (2016a:32): *"This was achieved by paying attention not only to how the algorithms work, but also how these calculative devices justify themselves as part of their framing the problem that they claim to be solving."* The key point of the research is that intentions induce choices, and choices get implemented as bias in code, and that this bias has profound consequences for how the same kind of (ranking) algorithm behaves. In general, algorithms *are* biased in many ways Bozdag (2013), for example by the data, by learning procedures, by programmers who make choices, by technological constraints and many other reasons. This immediately requires us to form an opinion about algorithms and whether they do *the right thing*, which again brings us back to ethical reasoning and consequences.

Ethical concerns about algorithms are a relatively new phenomenon and field of study. In the general media considerable attention is given to the topic in recent years. Open expressions of concerns by scientists such as Stephen Hawking and entrepreneurs such as Elon Musk and Bill Gates warn[100] for the unforeseen consequences of widespread use of AI. A letter[101] of concern with *"research priorities for robust and beneficial AI"* was quickly signed by more than 8000 researchers and practitioners. Individual top AI researchers speak out, such as Tom Dietterich[102]. Big tech companies such as Google, Amazon, IBM and Microsoft announced that they are forming an alliance[103] which *"aims to set societal and ethical best practice for AI research".* Various academic initiatives arise around the broad topic of "societal implications of algorithms", such as the newly founded ethics center[104] at Carnegie Mellon University (CMU). Numerous recent scientific papers warn for the many old and new ethical dilemmas that will arise when artificially intelligent algorithms are being employed throughout society (Mittelstadt et al. 2016).

In previous work (van Otterlo 2014a) I drew[105] on Walden Two, the fiction novel by the psychologist B.F. Skinner, to address the dangers of a *behaviorally conditioned society* that arises because of technological capabilities of algorithmization. A related warning came from Zuboff (2015) who defines the *"Big Other"* as a metaphor to point to the combined logic of capitalism, surveillance and digital technologies such as AI. Morozov[106] sees similar patterns of information capitalism undermining our human democracy. Shanahan (2015) tells a parable[107] of several AI systems working together and against each other in the society of the future[108], and makes clear that *multiple* superintelligent systems will make ethical challenges even more serious. These examples go beyond relatively simpler issues such as privacy and data protection, and see the potential influence of algorithms on society as a whole, with profound implications for democracy and free will. Many examples show that algorithms are not infallible, objective or trustworthy. For example, Google's search algorithm tagged[109] (photos of) black people as "gorillas", showing either a bias in data or learning procedures, or errors in the application of the tagging algorithm. Autonomously driving cars constantly make mistakes[110] or are not yet fully capable of driving in our complex, physical world. Even IBM's Watson algorithm, that won the typical "human" game Jeopardy, makes mistakes[111] too. Another related case is when algorithms are *deliberately* used for the wrong purposes. A good example is the fraud with testing software for cars running diesel fuel in recent years, the so-called Dieselgate[112]. Other examples where simple algorithms have large consequences are the mentioned Pokemon game, and the problems of tourists flooding big cities throughout the world because of the (algorithmic) services like AirBnB[113] and Uber[114].

---

99  https://arxiv.org/pdf/1702.05437.pdf
100 http://observer.com/2015/08/stephen-hawking-elon-musk-and-bill-gates-warn-about-artificial-intelligence/
101 https://futureoflife.org/ai-open-letter/
102 https://academic.oup.com/nsr/article/doi/10.1093/nsr/nwx045/3789514/Machine-learning-challenges-and-impact-an
103 https://www.theguardian.com/technology/2016/sep/28/google-facebook-amazon-ibm-microsoft-partnership-on-ai-tech-firms
104 https://www.nytimes.com/2016/11/02/technology/new-research-center-to-explore-ethics-of-artificial-intelligence.html?mcubz=1
105 See (in Dutch) this article for more "surveillance metaphors": https://decorrespondent.nl/1237/zo-kunnen-we-beter-over-surveillance-praten/41215603-67a22001
106 (In German) http://www.sueddeutsche.de/digital/alphabet-google-wird-allmaechtig-die-politik-schaut-

With new, digital technologies, legal developments are often too slow to capture a shift in moral values and novel possibilities, and *creepy* things arise (Tene and Polonetsky (2014): *"In certain cases, creepy behavior pushes against traditional social norms; in others it exposes a rift between the norms of engineers and marketing professionals and those of the public at large; and in yet others, social norms have yet to evolve to mediate a novel situation."* An example is the "girls-around-me"[115] app which uses FourSquare data to find, literally, the girls that are known to have checked in in your immediate surroundings. Nothing illegal is being done, although certainly the limits of decency and privacy are reached. Equally creepy are modern toys with sensor technology. The Cayla[116] doll caused some debate since it can communicate with children, send their data (voice, things said, possibly video capture) to the manufacturers' servers, and in addition, it can say anything to a child through a microphone. Apart from possible hacks, "connected" dolls are creepy because they invade the privacy of intimate family life, without doing anything illegal. Similar things can be said about smart-TVs and cell-phones that are connected too and equipped with many sensors.

### (5.2) A Taxonomy for the Ethics of Algorithms

As said, algorithms are not objective. Inside the code of an algorithmic system many *decisions* have been made which influence how the algorithm works, how it uses data, how it selects data, how it computes decisions, and how decisions affect the real world and real people. This also entails that ethical values are *built into* these algorithms, knowingly or unknowingly. A large literature is being formed around the *ethics of algorithms* in recent years (see for pointers: Lichocki et al. 2011, van Otterlo 2013,2014a,2014b,2016a, Medina 2015, Mittelstadt et al. 2016). The previous sections show that a huge diversity of examples of algorithmization exists, and also that there are many types of algorithms, ranging from simple sorting algorithms to the control structures of a superintelligent robot. Algorithms can be used for various *operations* on data, such as *classification* or *prediction*. Determining the actual (and potential) impact of such operations is difficult, for example because algorithms often operate in a larger context and because algorithms may

---

hilflos-zu-1.3579711

107 This story has many relations with a the current TV series "Person of Interest" http://www.imdb.com/title/tt1839578/

108 Predictions about the future (of technology) are notoriously hard. An interesting case are the predictions of the science fiction writer Asimov about our time, which are remarkably accurate, see: http://www.bbc.com/news/technology-27069716 and http://www.wired.co.uk/article/asimov-2014-technology-predictions

109 https://www.theverge.com/2015/7/1/8880363/google-apologizes-photos-app-tags-two-black-people-gorillas

110 https://phys.org/news/2016-09-dutch-police-probe-fatal-tesla.html

111 http://blog.chron.com/techblog/2011/02/jeopardy-watsons-more-interesting-when-it-fails/

112 https://www.cleanenergywire.org/factsheets/dieselgate-timeline-germanys-car-emissions-fraud-scandal

113 http://www.politico.eu/article/amsterdam-tourism-airbnb-crime-netherlands/

114 https://www.theguardian.com/cities/2016/oct/06/the-airbnb-effect-amsterdam-fairbnb-property-prices-communities

115 https://bits.blogs.nytimes.com/2012/03/30/girls-around-me-ios-app-takes-creepy-to-a-new-level/?mcubz=1

116 https://www.nytimes.com/2017/02/17/technology/cayla-talking-doll-hackers.html?mcubz=1

also adapt their functioning over a longer period. To better predict what are the possible implications of a particular type of algorithm, we need to look at the core dimensions of how they operate on data, in a non-technical way. In the following I will discuss two possible taxonomies that are useful for studying the impact, or ethics, of algorithms. The first is based on the *map* presented by Mittelstadt et al. (2016) which contains *concerns* about how algorithms transform data into decisions, which are then coupled with typical ethical issues. The second, discussed afterwards, is based on algorithm *capabilities,* and is complementary to the first and comes from a course I developed on this topic.

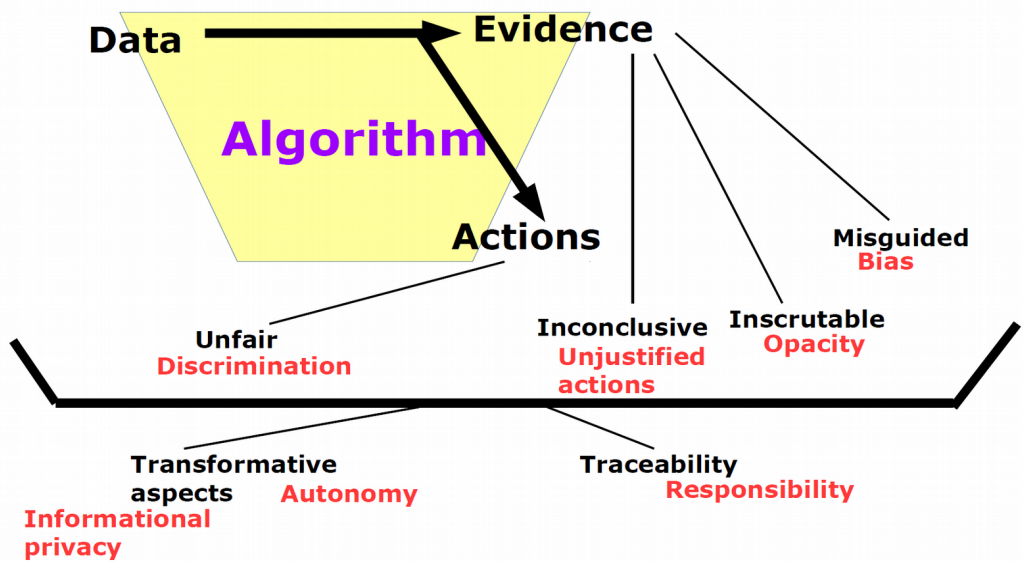

One useful way to order the ethical implications of algorithms is to look at how algorithms operate (Mittelstadt et al. 2016). In the figure I summarize the taxonomy; the blue area denotes the core operations of an algorithm: 1) it turns *data* into *evidence* which can be a probabilistic prediction, a yes-no decision, or some other *conclusion*, and 2) it uses the evidence to *trigger and motivate an action* based on the data. For example, an algorithm for bank loans could take personal data of someone and produce a credit-score of 12, which then could trigger an action to approve a particular mortgage. Because this action is taken in a large decision making context (e.g. bank regulations, a human operator, possible other decisions that are related, etc.) 3) the *apportionment of responsibility* (i.e. who is responsible for the overall *algorithmic decision making*) can be quite complicated. These three items form the basis of three types of *concerns* about algorithmic decision making: **Epistemic concerns** are about the transformation of data into evidence, and unfold in three forms. First, evidence may be **inconclusive** meaning that it may be probable, but inevitably uncertain. For example, when an algorithm predicts that I am a terrorist with 43.4 percent probability, how "exact" is that? Second, evidence may be **inscrutable** which means that the relation with the underlying data is not accessible for inspection, or incomprehensible. A good example are *no-fly lists* for suspected terrorists. Third, evidence can be **misguided** because the underlying data is incomplete or unreliable. Another type of concern is **normative** and is about (the consequences of) actions. One is that outcomes can be **unfair,** for example that they are discriminatory. Another normative concern is that algorithms can have **transformative effects**, which go beyond the issues when turning data into evidence, and are about how their operations fundamentally change how we perceive and understand the world. An example is profiling, which can *reontologize* the world and which can cause many changes in, for example, shopping and web use. A last concern is **traceability**, which is about the fact it can be hard to find out exactly how and why harm was done by an algorithmic decision. As a consequence, attributing **responsibility** is hard too.

These six concerns lead to a *map* of seven types of typical *patterns* of how algorithms cause new ethical challenges. First, inconclusive evidence may lead to **unjustified actions**, for example when someone is targeted as a terrorist on the basis of uncertain predictions. This can also be due to the use of (statistical) information of populations, applied to a particular individual. Second, inscrutable evidence may lead to **opacity**, meaning that actions based on such information cannot be contested, which also points to a *power imbalance* between the affected and the owner of the algorithm. Overall, *comprehensibility* of algorithmic decision making is often hard to ensure not only because details are often secret, but also because it is difficult to convey the full meaning of complex, statistical models in human-understandable form. A typical term for this incomprehensibility is saying that algorithms are *black* boxes. Third, misguided evidence may lead to **bias**. For example, if data is not *representative* then statistical algorithms will inevitably have a bias towards those that *are* represented in the data. Fourth, unfair outcomes may lead to **discrimination**. For example, when an algorithm pays attention to particular *attributes* of people in a database, it may discriminate indirectly, very similar to *redlining*[117]. Similarly, *personalization* algorithms segment populations such that only some groups get particular opportunities or information. Fifth, transformative effects may lead to **challenges for autonomy**. Personalization and filtering algorithms, and the resulting "filter bubbles" (Pariser 2011), may cause (e.g. for advertising revenue) that it is profitable to only show particular websites in the search results of particular individuals. This may withhold vital information that individuals need for decisions, and therefore has an impact of the individual's *autonomy*. Sixth, transformative effects may also lead to **challenges for informational privacy**. This is probably the most well-known ethical challenge and deals with many cases where users are not in control anymore of data about themselves. Last, traceability may lead to **moral responsibility**, which refers to the challenge of attributing responsibility when it is difficult to pinpoint the exact decision in the black boxes that algorithms are. In addition, *moral* responsibility may also refer to the ability to, simply said, *blame the algorithm itself*, which also requires attributing *moral values* to the algorithm and a capability to act upon them.

**(5.3) The Ethics of Agency in Algorithms**
A second taxonomy that I use is based the *level of agency*, or *autonomy,* we can attribute to an algorithm. In this taxonomy we simply look at *what the algorithm* can do, often in human-understandable form. This results in broad classes, see the figure, of algorithms that are partially ordered by "how complex" their operations are in human terms. Agency can be defined as *the capacity, condition, or state of acting or of exerting power.* The combination of the two taxonomies is useful to identify potential ethical challenges in the archival practice.

- Superintelligence
- Robots/IoT
- Optimization
- Learning
- Inference

**(5.3.1) Algorithms that Infer**
The first type consists of *algorithms that reason, infer and search.* These algorithms can be quite complex in what they do, but they all compute answers based on data *as it is*. The more complex they are, the more information they can extract from that data. Typical examples are database query algorithms, search engines on the web, and various algorithms for the interpretation of sound, text and images. The capabilities of algorithms to *extract* information from large amounts of data is vastly expanding every day, but is not new, and so is not the general threat to *privacy*. Long before the infamous "don't be evil"-slogan by Google, Warner and Stone (1970:146) already warned us to "not be naive about it": *"Anyone who has entered into a hire-purchase transaction, or any form of credit purchase, should nowadays expect both the personal data he supplied in his application, and the information about his reliability in making the repayments, to be widely available."* Furthermore, Warner and Stone (1970:137): *"At present we achieve some privacy because Big Brother has trouble scanning all his files at once. But, to the distress of the National Council for Civil Liberties and some computer engineers, there is a risk that computers may now make the bureaucrat's dream a reality and supply a complete dossier at the flick of a switch. There are such obvious medical advantages in this that it is likely to come about unless the public shows unusual vigilance."* Warner and Stone (1970:148) also note that *"Threats to privacy which are not commonly known or recognized as such thus already exist.... When all the records are computerized, and that proverbial 'press of a button' will release them, who can doubt that there will be greater and ever greater leakage and abuse?"* An additional problem is that privacy is often not well understood, and people often display differences between *intentions (or, values)* and *behavior.* Acquisti et al. (2015) performed many experiments with people in physical environments to test how much information people would share. It turns out that people show ambiguous privacy behavior because of the uncertainty in many situations, because of context switches (e.g. when mixing offline and online social situations), and because of outside manipulations (by companies and governments that are making profit with highly sophisticated tricks to lure people into giving up their privacy, using nudges, incentives or misleading information). Quoting the authors (p514): *"Insights from the*

---
117 https://en.wikipedia.org/wiki/Redlining

*social and behavioral empirical research on privacy reviewed here suggest that policy approaches that rely exclusively on informing or "empowering" the individual are unlikely to provide adequate protection against the risks posed by recent information technologies. Consider transparency and control, two principles conceived as necessary conditions for privacy protection. The research we highlighted shows that they may provide insufficient protections and even backfire when used apart from other principles of privacy protection."*

Algorithms nowadays, as opposed to the *database era* Warner and Stone talk about, do not only employ *pre-defined* database-items, but can interpret increasingly more formats that were once only possible for humans to understand, such as *images*, *sounds*, and *text*. For example, Google translate[118] now handles many languages quickly and efficiently and Google's algorithms can even deliver some "amazingly mournful" poetry[119]. More *semantic* understanding of text, for example *spatial[120]* expressions like *"the book is on the table next to the plant"* are being understood by algorithms. Generally, algorithms also become more sophisticated by incorporating knowledge about the world (e.g. by reading Wikipedia pages automatically) or by employing *how-to* knowledge[121], all typically in the area of the semantic web[122]. Lately, in the wake of *deep learning*, lots of progress has been made in interpreting *visual* information. For example in recognizing[123] *what is on a picture* in terms of objects and their relations, but also in evaluating the *aesthetics[124]* of a picture. The familiar Capcha puzzles[125] on the web highlight the importance of human visual recognition skills, but computers are quickly catching up. Also the *generation* of visual information, such as 3D models of faces[126] from pictures, and full *augmented reality* in which IKEA[127] furniture can be placed virtually in your own living room, real-time, are state-of-the-art examples. Google even helps its self-driving cars to recognize kids in Halloween[128] costumes as pedestrians as a visual skill. These recognition skills are entering the public space, such as cameras in billboards[129], recognizing types of people for marketing purposes. In terms of sound, algorithms are already better than humans in certain tasks such as speech recognition[130] and lip reading[131] and, for example, Skype conversations can now be translated on-the-fly[132]. General *data science* examples are abundant: for example Facebook can employ their data to *infer[133]* when people get into love relations. Still, algorithms have a long way to go to master all human-like skills of interpretation. For example, *deep understanding* of texts is still difficult, since it relies on commonsense reasoning and *relevance*, a concept that is not yet understood fully. Take the question *"Could a crocodile run a steeplechase?"*. The answer should be that it is impossible with short legs, but this requires quite a bit of reasoning over *relevant* knowledge. Also so-called *Winograd[134] schemes* are easy for humans, but not for algorithms. Let us take the sentence *"The city councilmen refused the demonstrators a permit because they [feared/advocated] violence"*. Depending on the verb chosen, the "they" refers to a different group, but such inferences are hard for (statistical) algorithms still. On the other hand, two decades ago McCartney and Anderson (1996) managed to formalize the logic of Roadrunner cartoons, which shows that in the end nothing will be impossible. The ethics of interpretation algorithms carries all *epistemic* and *normative* concerns about data, evidence and actions. However, now that even images and text can be understood, and linked with other data (for example faces with text and recent Twitter behavior), the *reach* of algorithms, and with that the impact of such algorithms especially on *privacy*, has become much more severe.

---

118 https://translate.google.com/?hl=nl
119 http://www.wired.co.uk/article/google-artificial-intelligence-poetry
120 https://www.wordseye.com/
121 http://www.wikihow.com/Main-Page
122 https://en.wikipedia.org/wiki/Semantic_Web
123 https://www.theverge.com/2017/6/15/15807096/google-mobile-ai-mobilenets-neural-networks
124 https://petapixel.com/2016/10/08/keegan-online-photo-coach-critiques-photos/
125 https://www.google.com/recaptcha/intro/android.html
http://www.captcha.net/
126 https://petapixel.com/2017/09/20/ai-tool-creates-3d-portrait-single-photo/
127 IKEA augmented reality https://www.youtube.com/watch?v=UudV1VdFtuQ
128 http://www.dailymail.co.uk/sciencetech/article-3301013/Google-teaches-self-driving-cars-drive-slowly-children-dressed-up.html
129 https://www.iamexpat.nl/lifestyle/lifestyle-news/hidden-cameras-dutch-advertisement-billboards-ns-train-stations-can-see-you
130 https://www.technologyreview.com/s/544651/baidus-deep-learning-system-rivals-people-at-speech-recognition/
131 https://www.technologyreview.com/s/602949/ai-has-beaten-humans-at-lip-reading/
132 https://futurism.com/skype-can-now-translate-your-voice-calls-into-10-different-languages-in-real-time/
133 https://www.facebook.com/notes/facebook-data-science/the-formation-of-love/10152064609253859/
134 https://en.wikipedia.org/wiki/Winograd_Schema_Challenge

A second type of inference algorithm is *search*. Algorithms such as Google's PageRank and Hummingbird[135] lookup billions of queries each day. And with the sketched advances in *semantics* and *understanding* texts and images, Google uses knowledge from many sources including Google Books[136] to aid its search algorithm. In fact, Google builds its own *knowledge graph* to make understand the queries its get better in terms of semantics (Vanderbilt 2013; Metz 2016). For example, the word "Ajax" can mean a soccer club, a programming language, or a Greek god. According to the motto *"things not strings"*, the context of use (information about the user, about previous searches, about difference in interest in the terms) can determine which one Google thinks is best, and which determines the specific search results. A simultaneous trend is that search engines more and more try to give *answers* instead of just links to webpages. One can now ask Google "how high is the Eiffel tower" and it will directly give the correct answer. Search engines also use increasing amounts of knowledge about the world and the user in the *ranking* of search results, but similar algorithms are used for Instagram feeds, or Facebook news feeds (*Edgerank*, see Birkbak and Carlsen 2016b). However, many users are not aware of the amount of *bias* in a search engine algorithm ranking their results (Bozdag, 2013). The automated calculation of this relevance ranking is on the one hand necessary for users in a world overloaded with information, and on the other hand always in danger of flattening the world to a point where the user is unable to be affected by it. Filtering information can also quickly become a (*transformative*) threat to *autonomy* because the search engine determines which information a user gets to see, and which not (van Otterlo 2016a). Also modern forms of *fake news* and *censorship* discussed in the introduction fall under this problem. And similar to how the search algorithm can be heavily biased, so can be the *indexing process*, which happens constantly when *webcrawlers* explore the internet to find material on webpages. The choice for when and where to do this indexing also defines what can be found (or is never to be found[137]) by typing in a search query, something which triggers all of the *epistemic concerns* about the underlying data. Much research has been devoted to this enormous *power* search engines have when it comes to information daily intake by people (Granka 2010, van Otterlo 2016a, Anthes 2016). Search engines are key *gatekeepers* in our society and influence the minds of billions of people everyday, usually without much oversight or democratic control, which has profound implications for *autonomy* again. They have been shown to be capable of influencing elections because of this (Anthes 2016, and see the *search engine manipulation effect* (SME)[138]), which is a serious ethical problem. The fact that these algorithms in addition, want to answer questions is even more ethically challenging. For example, it is nice to directly give the result of *"35 pounds in euros"* or *"opening hours of Albert Heijn in Nijmegen"*, but I would not like a search engine to (try) to answer questions like *"is it ok for Israel to close down the Palestinian areas?"* or *"is there a God?"*. Going in this direction can be called, in analogy with big data, *big knowledge* (van Otterlo 2016a). Next to search engines, other platforms are influential in similar gatekeeping ways; for example Facebook is the primary news source[139] for many people. Finally, because all information *empires* such as Facebook, Google and Twitter are *commercial* entities may mean that information about the data, or the specifics of algorithms, are *company secrets*. This, in turn, raises issues with opacity and inscrutable evidence, misguided evidence (and biased search results), but also with transparency and traceability of the overall algorithm.

Answering queries is an important issue outside the search engine community too. So-called *conversational agents* are algorithms designed to communicate with humans, in human-understandable form through text and speech. One form are the *social bots* (Ferrara et al. 2016) found on the internet, on social networks and in internet forums. Social bots can influence discussion on forums, they can act as genuine users on a platform such as Twitter (e.g. following other people, and reacting to posts) and so on. An ethical issue is that bots could be used for malicious purposes, such as steering a debate towards a particular outcome, or providing false support for election candidates. This raises threats for autonomy again as a transformative effect. Finding out which users are actual social bots is easier when a bot is clearly misbehaving such as Microsoft's Tay[140], but is generally a tricky[141] problem since it amounts to classifying it solely from its language use and electronic behavior, which will become more and more human-like since algorithms' capabilities are growing. A second type of conversational agent are the *voice-controlled assistants*[142] such as *Home, Cortana, Siri, Alexa* and several others. Such assistants can perform tasks like keeping keeping and creating shopping list, and *answering questions*. The main difference with search engines is that assistants will always answer the question because giving a list of links instead is impossible through speech, thereby aggravating the ethical problems concerning *autonomy of the user*. In addition, more information of the user is used (e.g. speech) which raises privacy issues again. Assistants like these are rapidly spreading through society, especially in China[143], and have already appeared[144] in legal[145] situations (as a "witness").

---

[135] http://searchengineland.com/library/google/hummingbird-google
[136] https://www.theguardian.com/books/2016/sep/28/google-swallows-11000-novels-to-improve-ais-conversation
[137] https://www.theguardian.com/technology/right-to-be-forgotten
[138] https://algorithmwatch.org/en/watching-the-watchers-epstein-and-robertsons-search-engine-manipulation-effect/
[139] https://www.theguardian.com/media/2016/jun/15/facebooks-news-publishers-reuters-institute-for-the-study-of-journalism

**(5.3.2) Algorithms that Learn**
The second class of algorithms goes beyond the first and can **learn**, and **find generalized patterns** in the data. These **inductive** algorithms perform statistical inference to derive patterns, models, rules, profiles, clusters and other *aggregated knowledge fragments* that allow for statistical predictions of properties that may not be explicitly in the data. Examples include algorithms that learn to classify images, and profiling algorithms. Overall, these are typically **adaptive** versions of the inference algorithms we have discussed, i.e. search engines typically adapt over time, and algorithms that interpret text, images and sound are often trained on such data. Applications range from predicting sounds for video[146], to training self-driving cars using video game data[147], even to predicting social security numbers[148].

Once algorithms start to *learn* (see Flach, 2012; Domingos 2012, 2015; Jones 2014; Jordan and Mitchell 2015) from data, many *epistemic* concerns come into play since the interplay with the data determines the algorithm's actions. Especially inconclusive evidence plays an important role since adaptive models are typically *statistical models* with uncertain outcomes. The bias comes in the form of a choice for a *model class* for which typically *parameters* are learned from data (van Otterlo 2013). In addition, bias is *emergent* (Bozdag 2013), in the sense that it is caused by the data and can change over time when the data distribution changes. Most powerful contemporary learning algorithms, such as *deep learning*[149,150], are purely statistical algorithms and very much like black boxes, which entails they are intransparent and the evidence they produce inscrutable (with some exceptions[151]). Other types of models are more *declarative* and more amenable to understanding and *explaining* decisions.

A typical area for ML is *profiling* and *personalization*, in which algorithms learn from user data how to adapt services, search results or recommendations to individual's tastes (van Otterlo 2013). De Hert and Lammerant (2016:146): *"In general, a profile is a set of characteristics, features and attributes with which a person or a group can be discerned from another person or group."* Facebook's news feeds but also Netflix's recommendations are driven by such algorithms, determining the user's choices and therefore affecting his *autonomy of choice*. An alternative is to let the user create his own profile by stating his preferences explicitly. If profiles are learned from data, algorithms typically learn statistical models *from many* users and

---

140 https://www.theverge.com/2016/3/24/11297050/tay-microsoft-chatbot-racist
141 As a matter of fact, it becomes similar to a Turing test (https://en.wikipedia.org/wiki/Turing_test)
142 http://www.businessinsider.com/siri-vs-google-assistant-cortana-alexa-2016-11?international=true&r=US&IR=T
143 https://www.technologyreview.com/s/608841/why-500-million-people-in-china-are-talking-to-this-ai/
144 See also the hilarious Southpark episode on these assistants: http://www.ibtimes.com/south-park-season-premiere-sets-amazon-echo-google-home-speakers-2590169
145 https://www.wired.com/2017/02/murder-case-tests-alexas-devotion-privacy/
146 https://www.engadget.com/2016/06/13/machines-can-generate-sound-effects-that-fool-humans/
147 https://www.youtube.com/watch?v=JGAIfWG2MQQ
148 https://www.wired.com/2009/07/predictingssn/
149 https://www.wired.com/2017/04/googles-dueling-neural-networks-spar-get-smarter-no-humans-required/
150 https://machinelearningmastery.com/inspirational-applications-deep-learning/
151 http://www.sciencemag.org/news/2017/07/how-ai-detectives-are-cracking-open-black-box-deep-learning

*apply them to a single user*. This may render inconclusive evidence which may be right *on average* but not for that single individual. Personalization algorithms do *social grouping* which create *opaque* contexts in which a person is *stereotyped* by how the algorithm sees that person, without sharing that knowledge. This induces many concerns related to privacy and discrimination, but also on autonomy. And although profiling algorithms are typically not concerned with exact identities (but: categories) it is a myth that anonymity is guaranteed when enough about a person is known, as has the famous AOL[152] case shown. A new privacy risk of profiling algorithms is that they can also reveal *new* knowledge (van Otterlo 2013). For example, algorithms can predict whether a person is *gay* or *muslim* from his Facebook *like's* (Kosinski et al. 2013) or whether a person is extravert based on *language use* (Schwartz et al. 2013). Computers can often predict personality traits better than colleagues, friends and family (Youyou et al. 2015) and eventually such traits may just be interpreted from a photo[153]. Such algorithms obviously have effects on privacy, but certainly also transformative effects related to *autonomy*. Moreover, all these data-related aspects raise issues on whether the data is *representative*, about which data was used for learning and how possible outcomes of the algorithm can be evaluated. Traceability in adaptive algorithms is an even more challenging problem.

In general, adaptive algorithms move society into a direction which can be characterized as *"the end of code"* (Tanz 2016). In the near future, increasingly many algorithmic decision making tasks will be learned from data, instead of hardcoded by programmers. Algorithmic models will be only *partially* set up (programmed) after which learning procedures fill in all the details. This has consequences for people, who will more often be assigned the role of *supervisor*, *parent*, or *trainer,* instead of *designer* or *programmer*. This has ethical consequences on a global scale, far more reaching than is contained in the 6 individual ethical concerns, since it will change how people work and how society functions. Technically, many interesting problems need to be solved too in order for human-algorithm teaching becomes feasible on a large scale[154].

**(5.3.3.) Algorithms that Optimize**
The third class of algorithms consists of algorithms that **optimize, incorporate feedback, and experiment**. These typically employ **reward functions** that represent what are *good outcomes,* which can be, for example, a sale in a webshop, or obtaining a new member on a social network. Reward functions tell an algorithm what is *important* to focus on. In general, algorithms do not deal with exact rewards, but rather compute *expected* rewards since there is a lot of *uncertainty* in any optimization problem. An example would be five throws with 2 dice: I do not know exactly which numbers I will get, but I can say that the expected total number of eyes will be more than 20. Optimization algorithms will then, based on all that is known about statistical aspects and based on all data about a problem, compute the *best* expected solution. Note that this combines aspects of *reasoning* and *learning* algorithms discussed previously. Earlier in this text I mentioned Deepmind's *AlphaGo* program which beat the best human player in the board game Go. AlphaGo is a typical example of an optimization program that combines reasoning, learning and optimization, based on the *reinforcement learning paradigm* (see Wiering and van Otterlo 2012), a subfield of ML for solving *multi-step decision task with uncertainty*. Reinforcement learning has a long tradition (see Menace[155] for an early example, and the origins in behavioral psychology and Skinnerboxes[156]) in AI, although recently interest has risen much (Krakovsky 2016).

The optimization setting features different kinds of rewards. In addition to taking the algorithm's point of view (e.g. sales, new members) we can also take the viewpoint of *users*. Take the Pokemon example from the beginning, in which the algorithm tries to optimize the rewards it gets by people playing the game and spending money. It does this, among other things, by *rewarding users* by giving them interesting creatures in a particular environment or by giving them other opportunities and items. This will *manipulate* (or: *nudge*[157]) the players in the right direction where the algorithm wants them since they help to optimize *its* expected rewards (or more precisely, that of the maker, Niantic Labs). This double optimization (algorithm rewards, and human behavior through rewards given to humans) is the basis for the manipulation of people's behavior. Most ethical concerns described matter in the case of optimization, but most prominent are transformative aspects. By optimizing over many users, final solutions are bound to be good *on average*, in a similar way as with learning algorithms. Optimal solutions for a *viral marketing campaign* could send mailings to clearly uninterested users, which would not be much of a problem, but it could also – in the case of a search engine – deprive particular users of particular content or advertisements. Even worse, it could nudge such users' behavior in the wrong direction, just for the sake of global optimization, which would also make the algorithm discriminating and unfair.

---

152 https://en.wikipedia.org/wiki/AOL_search_data_leak
153 https://www.theguardian.com/technology/2017/sep/12/artificial-intelligence-face-recognition-michal-kosinski
154 (MLIS 2015) http://proceedings.mlr.press/v43/
    (MLIS 2014) http://www.aaai.org/Library/Workshops/ws14-07.php
    (MLIS 2013) http://dblp.org/db/conf/ijcai/mlis2013
155 http://chalkdustmagazine.com/features/menace-machine-educable-noughts-crosses-engine/
156 https://en.wikipedia.org/wiki/Operant_conditioning_chamber
157 https://en.wikipedia.org/wiki/Behavioural_Insights_Team

In addition, optimization algorithms typically *iterate* the optimizations by **experimenting** with particular decisions, through *interactions with the problem*. A good example are algorithms that determine the advertisements on the web: they can "try out" (experiment) with various advertisements for individual users, and use the feedback (clicking behavior) of individuals to *optimize* advertisement placings. So, instead of a one-pass optimization, it becomes an *experimentation loop* in which data is collected, decisions are made, feedback and new is collected, and so on. In AI, this is a common technique. For example, the *Robot Scientist*[158] is a system that can autonomously do experiments (on yeast): inventing hypotheses, gathering data and testing hypotheses, analyses, and so on. Other physical examples are in robotics where experimentation is used to let a robot find out how interaction with physical objects works (Moldovan et al. 2012) and in IoT where experimentation is envisioned to explore how to optimize the number of books checked out in a public library (van Otterlo 2016b). Most experimentation happens in online worlds though since the marginal costs are small. Platforms with large user bases are ideal laboratories for experimentation. For example, Netflix (Gomez-Uribe and Hunt 2015) experiments with user suggestions to optimize their rewards which are related to how much is being watched. They combine personalized suggestions with fairly random ones, just to see "what happens". Facebook does many experiments with users, often without them knowing it. For example, they changed peoples moods and voting behaviors for millions of people at once[159]. Some creative counter-moves have been performed, for example Warzel (2016) showed what happens if you follow all suggestions by Facebook, and Honan (2014) did a hilarious Facebook experiment by "liking" everything that is available. Both experiments showed the influence algorithm has on what one gets to see on Facebook, but they also show how "brittle" the algorithms are when the user displays extreme behavior. Experimentation with data and algorithms nowadays happens for supermarket prizes, on the energy market, with airline tickets, website variants (Google A/B-testing) and for many other things.

A typical outcome of optimization is a *ranking* (e.g. viable options, users, documents). Often based on *feedback*, for example given by customers, an overall *score* can be computed and *compared against* scores of others. In the *ranked society* in which we now live everything gets ranked, with examples such as Yelp, Amazon, Facebook (likes), Tripadvisor, Tinder (swiping), OkCupid, all to find "the best" restaurant, lover, holiday trip, or book. Also in our work life, ranking and scoring becomes the norm (called: *workplace monitoring*[160]). The ultimate example is China's 2020 plan (Chin and Wong 2016) to rank everyone in society to find out *"how good a citizen are you"*. Scores are computed from many things ranging from school results to behavior on social media, to credit score, and combined into one overall score. The higher that score, the more privileges the citizen gets (from easier car rental and bank loans, to visa to other countries).

There are many ethical issues concerning optimization and experimentation. All concerns fire in this case. Again the most prominent ones are transformative effects on autonomy. By letting algorithms manipulate our information environment, by letting them nudge us with rewards, and by being ranked, people lose autonomy. Furthermore, they will resist, and try to find out *how* they are being manipulated and try to figure out how to *beat the system*, with examples how Uber drivers try to get the best rides by trying to figure out how the Uber algorithm *manages them* (Lee et al. 2015) or how people try to *rationalize* Facebook algorithm decisions about news feeds when (deliberately manipulated) information does not seem to make sense (Eslami et al. 2015). Being ranked or manipulated by algorithms is one thing, but suddenly changing rankings and rules trigger concerns about *fairness* and *traceability* of the system, with examples of Instagram's new ordering[161] algorithm freaking people out, and restaurant ranking sites' new rules creating unfair practices[162]. Puschmann and Bozdag (2014) discuss many issues (pro and con) in experimentation and stress the *ubiquity* and that they are often also beneficial for individuals. Overall, there are two main ethical issues that play a role here. One is the fact that in many cases we do not know, or do not want, that we are involved in algorithmic management or experiments: do we need new forms of *informed consent* in the digital world? A second issue is the reward function: who defines it and who, thus, has the ultimate power of deciding what is good, and what is bad, in a utilitarian sense.

---

158 https://en.wikipedia.org/wiki/Robot_Scientist
159 https://www.wired.com/2017/05/welcome-next-phase-facebook-backlash/
160 https://harpers.org/archive/2015/03/the-spy-who-fired-me/

**(5.3.4.) Physical Manifestations**
A fourth class of algorithms concerns **physical manifestations** such as **robots** and **sensors** (internet-of-things). These algorithms go beyond the digital world and have *physical presence* in our physical world. This also means that they can perform *physical action* which may jeopardize human safety.

A first manifestation is the *internet-of-things* (Ng and Wakenshaw 2017) in which many appliances and gadgets get *connected* and where increasingly so *sensors* are being placed everywhere, creating data traces of once physical activities. The *programmable world* (Wasik 2013) will feature all digital (and intelligent) items around us as being *one giant computer* (or: algorithm) that can assist us and manipulate us. For example, if your car and refrigerator and microwave could work together, they could – with the right predictions on the weather, your driving mood and speed, and possible traffic jams – have your diner perfectly cooked and warm the moment you get home from work. The ubiquity of such sensor-abundant systems comes with opportunities, but also with threats (see also van Otterlo 2016b and examples in there). The strongest one is that ubiquity means that it happens without being noticeable, such that loss of autonomy or privacy may come without warning. Since sensors are meant to gather data, and devices with agency to act upon that data, all mentioned concerns apply. In very limited time, devices all over the world will communicate more information than we humans, which has consequences.

A similar big development will be physical *robots* in our society. The word "robot" comes from the Chech word[163] "robotnik" which refers to serf, farmer, "boring work" and "hard labour". The difference between very early (AI) robots like Shakey[164] and the very modern, naturally walking, animal-like robots by Boston Dynamics[165] is huge but the core ingredients are the same: a physical manifestation of an algorithm able to act. Or more precisely: *"a robot is a constructed system that displays both physical and mental agency, but is not alive in the biological sense"* (Richards and Smart 2016). There are many types of robots currently, ranging from simulations, to humanoids and mobile manipulators[166], which are mobile robots that have arms and hands to carry out action on objects around them. For the latter, research activity is high in recent years (e.g. see Moldovan et al. 2012) because they are useful for various applications ranging from manufacturing to healthcare and to companion robots. In the library setting, several manipulation-capable robots have already been employed (Bdiwi and Suchy 2012; Li et al. 2013). Note that currently only very few robots are actually in people's homes, and the ones that are, are vacuum cleaners[167].

The promise (and safe assumption) in robotics is that one day they will be part of our daily[168] lives, but many technical advances are still needed to make it so. However, we can start to think about ethical consequences (Lichocki et al. 2011) before that happens. Steinert (2014) frames the ethics of robots into four main[169] categories: robots as *tools* (or instruments), robots as *recipients of moral behavior*, robots as *moral actors*, and robots *as part of society*. The difference between the first and the latter two is mainly one of *responsibility*. If we see robots as (possibly harmful) tools they are on par with knives and chainsaws, and even if we act morally towards them because we generally "feel" for them (which happens often in the interaction with humans, such as with office robots[170], but also with simple[171] robots) they are still "machines" with human operators who are responsible when something happens. In other words, we should not fall into the *android fallacy* which is about our default reaction to attribute too many (mental) capabilities to robots that look like "us" (Smart and Richards 2016). Such robots can, however, trigger other novel ethical dilemmas through their appearance and how life-like[172] (or even human-like[173]) they are. Only if we really enter the phase in which robots become *moral actors* (the third type) we also need to think about the ethical implications of *agents* which are not human (and presumably very smart, see the next section). Naturally, the internet-of-things and robotics will blend into one giant computerized system around us humans.

---

161 https://socialmediaweek.org/blog/2017/07/lurkers-on-top/
162 https://www.trouw.nl/home/horeca-klaagt-over-vals-spel-bij-iens~ae080a6b/
163 https://en.wikipedia.org/wiki/R.U.R.
164 https://en.wikipedia.org/wiki/Shakey_the_robot
165 https://www.bostondynamics.com/
166 http://www.mobilemanipulation.org/
167 http://www.irobot.com/For-the-Home/Vacuuming/Roomba.aspx
168 Many good movies exist about how that would look like (See: I Robot, the Bicentennial man, A.I., and Robot and Frank, for some examples). An excellent, fairly realistic, series is "Real Humans" (http://www.imdb.com/title/tt2180271/)
169 The article also includes a fifth type which refers to the *influence* of robots on ethics itself (meta-ethics).

The introduction of increasing numbers of robotic agent in society (the fourth type) will also have socio-economic consequences we can only partially imagine. Most obviously for *work* which will[174] increasingly being taken (or not[175]) over by robots (Ford 2013) and will require new ways to distribute income over humans. Jobs that can be broken down into predictable, routine-like tasks are in danger, including lawyers, paralegals, journalism and retail. It is hard to predict which jobs will stay "human" and which not, and also which new jobs (e.g. robot trainer) will arise. Apart from the job market, robots are expected to have profound (ethical) impact on many aspects of society, such law enforcement (Sharkey 2009) and military operations, sex and relations or even robot prostitution (Richardson 2016), autonomous cars and traffic (Kirkpatrick 2015), elderly care and healthcare, and many other fields.

### (5.3.5.) Superintelligence

The fifth class of algorithms goes beyond the algorithms as we know them now (digital or in physical form) all the way to **superintelligent** algorithms, which surpass our human-level intelligence. Once we have reached that point, questions of *conscience* and *moral decisions* and with that **responsibility** of algorithms will play a role. Most of this discussion falls beyond the scope of this text. A general remark is that the more intelligent, autonomous or conscience an algorithm will become, the more moral values will be attributed to it, and the more ethical reasoning and behavior will be expected of it. However, as (Richards and Smart 2016) elegantly show using the *android fallacy* it will take still a long time before robots are even capable of deserving that. According to many scholars, a so-called (*technological*) *singularity* (Vinge, 1993; Shanahan 2015) will come, which is[176] *"the hypothesis that the invention of artificial superintelligence will abruptly trigger runaway technological growth, resulting in unfathomable changes to human civilization"*. For some already the point of getting algorithms to become "smarter" than humans (whatever that may mean) will trigger an explosion of unstoppable AI growth that could dominate the human race entirely even. Ethical concerns about such algorithms are discussed by Bostrom and Yudkowsky (2011) and many other people, like Kurzweil[177]. Many straightforward ethical concerns are about whether machines will overpower us, whether they still need "us", and what it *means* to be human in a society dominated by machines (see Shanahan 2015 for some pointers). Such concerns about unbounded AI systems can trigger old ideas of the "emergency button" as discussed in the first section. A very interesting contribution to this discussion was given by Hugo de Garis[178] who thinks that the main coming battle will not be between humans and AI, but between humans who *want* to develop superintelligent AI and people who oppose it.

### (6) Towards the Ethical Codementalist

In the previous sections I have sketched how digitalization and algorithms can cause various ethical challenges. The *superintelligence* aspect, where AI can spin out of control relative to our human interests, is one that triggers *existential fears*. Such AI can be seen as a technology that can profoundly change the world. It reminds of concerns when another such technology was in its initial phases: *nuclear technology*. Albert Einstein warned president Roosevelt in 1939 in a letter[179] for the consequences if *some other nation* (Germany) would obtain the technology for powerful bombs and suggested to start a nuclear program in the United States. The current explosion in AI technology may very well trigger a similar arms race. But before worrying about superintelligence, we should be concerned about the many ethical challenges of not-yet-fully-superintelligent algorithms. Some functionalities of AI systems will appear superintelligent (for example having access to billions of databases in an instant) though. Human-level and superintelligent beings are not just a theoretical exercise, but something to start thinking about.

---

[170] http://snackbot.org/about-public.html
[171] http://www.mykeepon.com/
[172] https://en.wikipedia.org/wiki/Uncanny_valley
[173] https://www.wired.com/2017/04/robots-arent-human-make/
[174] https://www.wired.com/brandlab/2015/04/rise-machines-future-lots-robots-jobs-humans/
[175] https://www.wired.com/2017/08/robots-will-not-take-your-job/
[176] https://en.wikipedia.org/wiki/Technological_singularity
[177] https://en.wikipedia.org/wiki/The_Singularity_Is_Near
[178] https://www.forbes.com/2009/06/18/cosmist-terran-cyborgist-opinions-contributors-artificial-intelligence-09-hugo-de-garis.html

In this section, I want to deal with the implications of algorithms slowly taking over professions such as of the archivist or the librarian, by sketching a *constructive* approach to ensure humanly ethical behavior in such systems. What if we would bring the pioneer's dreams from last century, such as Otlet's Mundotheque, or Bush's Memex, or Wells' World Brain to the AI age, and what would it take for these systems to satisfy human ethical requirements? In order to make things more general, I invoke Otlet's concept of a **documentalist**, a profession covering many aspects of librarians, archivists and other information workers. (Wright:97): "*It would call for a new breed of professional, what Otlet called a "documentalist." Unlike the traditional librarian - whose task consisted primarily of collecting, archiving, and curating books - the documentalist would play a far more active role in analyzing and dstributing recorded knowledge. Applying Otlet's monographic principle and following the cataloging rules of the UDC, the documentalist would collect, analyze, and summarize documents from multiple sources, then disseminate them into a larger apparatus of recorded knowledge.*" In the remainder of this section I will appeal to documentalists to represent archivists, librarians and the like.

**(6.1) The Documentalist Singularity**
*It took ages to get permission, but yesterday evening I finally got THE mail. I consider myself lucky, since I really needed access to the archives to finish my article. Other people would ask why would an assistant professor in technology ethics would like to see those old-fashioned paper documents about the introduction of Mindbook, the company that grew out of the long gone Facebook corporation. Since, their summaries are already on Archipedia. Who is interested in paper documents anymore? Well I am. I never felt comfortable with all this digital.. eh stuff.. anyway. People are physical, and they like physical things. Well.. at least that's my opinion. And besides… I don't trust Archipedia; they have appeared in so many algorithmic trials for information manipulation, but they always use their right-to-silence and nobody is been able to crack their summarization code. I need to take a look myself. I enter the red building next to the rocket station and turn right after getting through the bio-scanner. Paul, a robot from the CODEMENTALIST-5000 series is waiting at the desk. His emotional module can use an update, I catch myself thinking. I only get a nod and a metallic "hello, how can I help you?", so much unlike the newer models that can really lighten your day with their big smiles and warm voices. I answer the way I am supposed to do, with a clear question and context: "Hello Paul, I'd like to see all documents containing discussions on the use of advanced mind models, especially whole brain simulations, of Facebook users prior to the formation of Mindbook. I also would like to look at pictures and footage of the meetings that include people from the legal department, and can you please provide me with additional CV information of these people? Thank you." Paul knew from prior contact that I would be coming to the archive myself; otherwise he would have downloaded the interpreted documents, or DOC-INTERPRETs as they call them here, to my communicator. Now he only sends the requested CVs and projects an interactive map of the archive a floor below which will guide me to the right boxes. Since Paul scans and stores all items (including photos and a shallow semantic analysis of texts), and organizes them in the physical space, he knows where I have to go. At least, that is what I have to believe since there is no way of knowing what is in the complete archive. While going downstairs, I sense excitement from my side on how optimized and effective my routing past all the right boxes, 16 in total, is. Five more boxes are off limits for me though. It turns out another researcher has a similar research question in parallel, and his (or her?) combined scientific h-index and social media coverage is so much higher than mine. Also, according to an analysis of the planned social activities in our agendas, and our biophysical energy levels in combination with the predicted moist weather in the next weeks, Paul estimates that I will*

---

179 https://en.wikipedia.org/wiki/Einstein%E2%80%93Szil%C3%A1rd_letter

*not put enough hours in my analysis of the documents and my writing anyway. Sure… I need to stop eating snacks and boost my metabolism… but come on… who does Paul think he is? My doctor? According to Paul the overall estimated impact of the other researcher publishing the material alone is higher when I do not interfere. I have no other option than to accept, but I don't think it's fair. Archival robots such as Paul are built to optimize their impact since they too get ranked. Of course, everyone gets ranked, and so are archival robots. Paul needs to optimize the use and costs of the archive while at the same time striking a balance between preventing possible negative impact on the donor organization Mindbook, and stimulating positive impact from researchers and journalists publishing the right kind of information, again according to Mindbook. Oh well… the rest of the day I look at the documents, trying to find what I am looking for. The surveillance-sensors watch my every move while interacting with the documents, which helps them to further optimize the archive, so they say. Well..they sure also use them for the projected advertisements that are appearing on the electronic walls for me. Hey… yes indeed… I do need a snack… my hands are trembling…. How did they know? Oh… never mind.*

Based on the previous sections, this scenario seems plausible. A robotic documentalist, or **codementalist** as I call it, represents a modern version of the pioneers' dreams (Memex, Mundotheque) described in the beginning of this text. "Codementalist" combines the characteristics of the documentalist with "code", the substrate of algorithms. It remains to be seen how the future will turn out, how far off this scenario is, and how long human documentalists work together with codementalists. What we can say is that algorithmic technology will have a huge impact on practice, and on ethics. Ferguson et al. (2016:549): *"It seems likely that, when considered in the abstract, new technologies do not appear to change ethical principles; however, when actually experienced in the workplace, they do substantially change the factors the … professional has to weigh up."*

In terms of digitalization, we can assume *everything* will be digitalized: the collection and all interactions between users, documentalist and collection. Access will be easier, but also less privacy-friendly. In addition, an abundance of data will be available on users, with links to other users, the collection and any piece of information available through digital means. This also means that *performance reports*, on collection use and reach, storage efficiency and services can be monitored in real-time and evaluated and optimized when needed. Rankings of users, documentalists and codementalist s will be kept on many scales. For example, archival behavior can be benchmarked, but also users can be compared to aid in ethical dilemmas of fair access. Inspired by China's 2020 ranked society, one can ask *"how good an documentalist are you?"* and connect that to privileges or wages for documentalist.

Due to algorithmization many possibilities arise to connect (potential) users to collections, to personalize user interactions and to optimize various aspects ranging from space usage (for physical parts of the collection) to advanced retrieval methods based on complex user demands. Codementalist s could also come up with new "business models", for example by sending personalized advertisements for parts of a collection, or by assembling user-targeted image aggregations and summaries for particular topics the user is (presumed to be) working on. In principle, one could try to turn typical marketing around, and try to find *the most probable person interested in this particular collection*.

These are just some opportunities. Many others await and Section 5 showed each comes with their own set of ethical challenges for individual users and society. It is not a matter *if* all this will happen, but *when*. I define the coming **documentalist singularity** as the moment when all core documentalist's activities will be replaced by an codementalist. Note that I am not talking about a general superintelligence, but a more specific singularity for the documentalist profession. Just like in autonomous cars, we can talk about various *levels*[180] of autonomous codementalist: some will only maintain digital archives, some will have a robot body (for physical collections), and some may only function as an assistant of a human documentalist. The main point I want to make is that since "we", as humans" are creating these codementalist, we may study their ethical implications before, during and after creation, but maybe much better is to try to *create them such that we can ensure that they will behave according to our own moral values*. How can we do that?

**(6.2) Some Solutions**

---

180 https://en.wikipedia.org/wiki/Autonomous_car#Levels_of_driving_automation

Algorithmic versions of virtually all current professions will appear, eventually. The basic, *human*, question is how to ensure that all these algorithms respect our human values and for this we should not rely on corporations. For example, the founders of Google, Brin and Page, promised at the start never to show adds with search results (broken in 2002). Google would never combine data from different sources (Youtube, calendar, gmail; broken in 2014). Google would not automatically combine data gathered by its *Doubleclick* company with other data – users needed to give permission (broken in June 2016). Examples for other services exist and do illustrate self-regulation will not be the final solution.

Current literature on ethics of algorithms often deals with how to (legally) *govern* algorithms in such a way that they *behave* appropriately (e.g. Etzioni and Etzioni 2016; van der Sloot et al. 2016). Diakopolous (2016) discusses several general ways to ensure *accountability* in algorithmic decision making and comes up with an *algorithmic transparency standard* which calls for *human involvement* and transparency about the data and (inference in) the models used. Although very desirable, and very often mentioned in other literature (see van Otterlo 2013), from the previous sections we can say that this will not be easy in most cases. For starters because private companies will not disclose all their software secrets (for example Facebook and Google do not disclose their exact ranking methods), and even if they would do, it would be hard to convey the meaning of enormous, and technical, models to the individual user who is affected by an algorithmic decision. Diakopoulous (2016) addresses this by calling for the *design of effective user experiences for transparency information, visualization to succinctly communicate the workings of an algorithm,* and *developing ML methods that can be explained in ways that humans can readily understand.* Probably the most effective way he proposes is to *make algorithmic presence known to the user* at all times.

A challenge is that so far algorithms are largely *unregulated*. However, there are laws and rules for *data*, such as the *data protection act* (DPA; Dutch: AVG[181]) from 1998. The DPA can be used when, for example, data is gathered in one place, say license plates scanned for checking parking permits, and used in another, for example to check where somebody has been with his lease car to see whether he drives more than allowed with that car. In that case data is used for a different purpose than when gathered and this is protected by law. Another example would be a loss of a USB-stick containing patient files from a hospital; in that case an even more protected set of *personal* data, namely *medical* data, is lost, which would amount to a severe privacy violation. In the Netherlands since 2016 it is required by law to report[182] any *data leak*. In Europe, new legislation is now being developed as an extension[183] to the DPA in the form of the *general data protection regulation* (GDPR[184]) which will cover several forms of algorithmic decision making. Individuals will have the right to be informed about it, the right to access (and erase) their acquired data, the right to let their data be transferred to another service provider, and several rights in relation to algorithmic decision making and profiling in particular. Whether the GDPR will be effective will depend on its practical application in difficult cases (see the discussion in Mittelstadt et al. 2016).

Outside the law, many other solution have been proposed, such as methodologies where protections are *built-in* (such as *privacy-by-design*, and *transparency-enabling technologies*) or where (data) protection is ensured by technologies based on *encryption* and novel technologies such as *blockchain*. Individual users can often protect their privacy to some extent by using software or services from companies that are privacy-friendly or more transparent, e.g. the privacy-friendly *Duckduckgo* search engine, *Whatsapp* alternatives such as *Signal* or *Telegram*, or services that guarantee data encryption such as on the *iPhone*. Yet other solutions to deal with *power imbalances* between big corporations and individuals propose to *monetize* data: trade privacy for monetary rewards. Tene and Polonetsky (2014) propose several solutions that ask for transparency and new (social) norms for technology (such as Facebook sharing behaviors). They are also *against targeting the superuser*: assuming that all users can just protect themselves. A solution shared by many is *data minimization* (see e.g. Medina 2015)*:* only gather data that is really necessary, which is often in conflict with companies' business models. Another set of solutions is *obfuscation* (Brunton and Nissenbaum 2013) in which users deliberately sabotage algorithmic systems, by for example submitting random search

---

181 https://www.autoriteitpersoonsgegevens.nl/nl/onderwerpen/europese-privacywetgeving/algemene-verordening-gegevensbescherming
182 https://autoriteitpersoonsgegevens.nl/nl/onderwerpen/beveiliging/meldplicht-datalekken
183 https://ico.org.uk/for-organisations/data-protection-reform/overview-of-the-gdpr/
184 General Dat Protection Regulation (GDPR) http://www.eugdpr.org/more-resources-1.html

queries mixed with genuine queries to prevent the search engine from profiling the user reliably.

Another way to tackle the ethical issues of algorithms is to *employ AI itself*. That is, one can utilize the same power of algorithms to deal with ethical issues. For example, recent advances in ML remove discriminatory biases by adapting training methods, or implement *privacy-aware* techniques. Other developments are algorithms that use the same kind of ML techniques as search engines do for profiling, to *counter* the search engine by generating *maximally confusing* search queries. In addition to the many specialized algorithms now being developed, Etzioni and Etzioni (2016) propose general, so-called *AI Guardians* to help us cope with the government algorithms. Since AI systems more and more become *opaque* (black box), *adaptive* (using ML) and *autonomous,* it becomes undoable *for humans* to check what they are doing. AI systems can do that for us, at higher speeds, with much more processing power, and working 24/7. AI guardians are *oversight* systems using AI technology, and come in various sorts: *interrogators and auditors* which can investigate or audit after e.g. a drone crash, or when unwanted things happen in very large software systems such as Google or Facebook. AI guardians can also *monitor* ungoing processes, or even *enforce* compliance with the law. A special type is the *ethics bot* which is concerned with ensuring that the operational AI systems *obey ethical norms.* These norms can be set by the individual, but can also come from a community. An ethics bot could guide another operational AI system, for example to ensure an autonomous car is not driving fast, or that a financial AI system only invests in socially responsible corporations. Ethics bots will also have to *learn* the moral preferences of their individual, either by explicit wishes or from observed behavior. The idea of using AI to derive what we can call *machine ethics* (Anderson and Anderson 2007), to ensure moral behavior of AI is something I will take further here.

**(6.3) Learning to Behave Well**
The most prominent AI area of this moment is ML. A promising idea is to let systems *learn* how to behave appropriately. One can imagine a codementalist watching a human documentalist doing his or her job and learning instantly what is the right way to act. Would it be that simple?

Let us look at a domain where machine ethics is a hot topic: *autonomous cars* (Kirkpatrick 2015; Goodall 2016; Bonnefon et al. 2016; Sütfeld et al. 2017). Autonomous cars are excellent examples where moral reasoning is require in physical worlds with matters of life and death. In case the car AI predicts an upcoming accident in the instant, it needs to decide quickly: kill the passenger by running into a wall, or kill the pedestrian (with child) by avoiding the wall. More extremely, the car AI could estimate life expectancies from appearance combined with statistical databases to make decisions even more finegrained. There are many good arguments why such situations *will* occur (Goodall, 2014) and somewhere in the car's software such decisions *will* be represented. ML is used to learn many technical aspects of *driving* itself, ranging from learning to recognize pedestrians to learning to take over other cars. However, learning how to *drive morally* is a different problem, but could in principle be learned from good examples too.

Since humans are currently the only drivers on our roads, it makes sense to investigate how they look at such problems. Bonnefon et al. (2016) shows that humans typically approve of *utilitarian* cars which look at the possible outcomes and choose the "best" moral decision. However, they also show that people prefer cars that would protect them as a driver, and would react negatively to government regulations that would prescribe general utilitarian decisions (where the driver could be sacrificed too). Sütfeld et al. (2017) go even further and put people in simulated driving (in *virtual reality*) experiments to see which decisions they would make when confronted with accidents involving people from different ages, balls, pylons, trash cans, dogs and more. Their findings show that a simple one-dimensional *"value-of-life"* scale (to compare the utility of objects) works. Another finding is that human decision-making becomes less consistent with time pressure.

This suggests that for clear choice problems between "victims" human moral behavior could be learned from examples. There are many technical ways to do that. As an example consider the system by Moldovan et al. (2012) in which a robot needs to learn, from humans, how to perform a skill such as *"reorganize a shelf with objects".* At first, the robot starts playing with objects to see how they behave when being manipulated by the robot's hands. In this phase, the robot *experiments* by trying out different *actions* and *seeing* their *results*, and *learns* the dynamics of the physical world. In a second phase, the robot is shown by a human how to reorganize several objects on a shelf. By *observing* the human and the *effects* on the objects, it learns which *configurations* of objects are *desired* by the human. In the final phase then, when faced with a new problem, the robot utilizes all this knowledge and *optimizes* his course of action to reach a desired *goal* state. Different phases may require learning different things: exactly replicating *actions*, replicating demonstrated *outcomes*, or obtaining a generalized model of general characteristics of "good" outcomes. An codementalist could learn the literal actions to reorganize parts of the collection, or it could try to learn *general principles*[185] which are reflected in the demonstrations of an documentalist such that it could reorganize other parts of the collection too.

An alternative to acquire *imitation* behavior is to learn the underlying *reward function* according to which the original behavior was generated. If that reward function is obtained, one can simply behave towards results that have a high value, instead of learning the exact action for each situation. Since moral values can, assumably, be equated to reward values, it gives an opportunity to learn moral behavior from observing a demonstrator. The technical term for this type of learning is *inverse reinforcement learning* (Wiering and van Otterlo 2012; Abel et al. 2016) which is based on solid theories for behavior learning. For specialized tasks, especially in robotics, many successful applications exist. This could form the basis for AI systems that act in alignment with human goals and values, which is an interesting option for ethical codementalist. The core challenge then is how to learn these *human* values, sometimes framed as *the value learning problem* (Soares 2015), or *value alignment (*Taylor et al. 2017), currently a hot topic in ethics and ML. The challenges is that human values are typically difficult to learn, since they can be based on complex mental processes, can be working on multiple timescales, can be difficult to put on one scale (although the autonomous car study suggests that sometimes that is possible), can involve both intuition and reasoning and may involve *signaling* to establish trust (Kuipers 2016). Furthermore they require *ontological agreement* between human and machine: do they see the world in the same way? Many of these problems are shared with technical AI work (e.g. computer vision) but for use in ethical systems much more work is needed. Taylor et al. (2017) list 8 grand directions to work on, all aiming at new ML technologies which can, for example, i) learn human values, ii) signal human operators when there is ambiguity in the data, iii) aid a human overseer, iv) optimize "mildly" (not trying too hard when goals have been "pretty well" achieved) and many more.

Reward functions are important, but not the final answer. AI systems can often find *non-human* ways to achieve the *human* values. Shanahan (2015:210) describes an AI tasked with maximizing the production of *paperclips* in a small manufacturing company, which seems like a simple reward function. However, this *superintelligent* AI ends up colonizing the whole planet for resources for making ever more paperclips which is not aligned with the original, simpler, intention of the task. Similarly, a police AI tasked with bringing down crime rates could just come up with a solution in which all citizens need to stay in their homes forever. AI systems can often, just like evolution[186], come up with surprising solutions to maximize a reward function and this is another opportunity to be *not* aligned with human values. A general problem with any adaptive ethical algorithm is that it can always diverge to unethical behavior over time.

**(6.4) Making Intentions Explicit**
The value learning problem is difficult for many reasons. In addition, any type of purely statistical learning procedure faces other difficulties related to opacity and the limited possibilities to employ *knowledge* one might have about a domain. In the previous section I have tacitly assumed that everything is learned from scratch, which is not optimal if domain knowledge is available. Take again the robot example from Moldovan et al. (2012), where part of the learning task is about learning statistics about object interactions. Here, additional knowledge can be added to make the learning task simpler. For example, one can specify that one is only interested in *how an action on one object changes the distance between that and another into far, medium or close.* Such prior knowledge specifies a *bias* on the learning process and makes it relatively easier but also more *biased*. Other types of knowledge for this task could be that *green objects are typically*

---

185 The real challenge is often generalization: "We're always learning from experience by seeing some examples and then applying them to situations that we've never seen before. A single frightening growl or bark may lead a baby to fear all dogs of similar size – or, even animals of every kind. How do we make generalizations from fragmentary bits of evidence? A dog of mine was once hit by a car, and it never went down the same street again – but it never stopped chasing cars on other streets." (Minsky, 1985, Society of Mind, Section 19.8).
186 See the intriguing experiments by Karl Sims:  http://www.karlsims.com/evolved-virtual-creatures.html

*heavy*, and *one cannot place an object on a ball-shaped object*. Various ML systems exist where this knowledge bias is *declarative*, meaning that the knowledge is specified in a special *language* or *database* and where this knowledge can be analyzed, processed, injected into a problem, and most importantly, *extracted after learning*. In addition, such methods support *reasoning* over that knowledge. Many **knowledge-rich ML** methods exist for use in behavior learning contexts (see van Otterlo 2009, 2013) such as the codementalist profession we are looking at. Furthermore, some declarative ML methods have been employed for ethical reasoning in AI (Anderson and Anderson 2007) and other ethical studies (van Otterlo 2014a).

A key issue is ontological: if we would like to employ declarative knowledge from humans, it should be at the right level and meanings should mean the "same" for the AI and for humans. To bridge AI and human (cognitive) thinking, the *rational agent* view is a suitable common view. In AI, a rational agent is *"one that acts so as to achieve the best outcome or, when there is uncertainty, the best expected outcome"* (Russell and Norvig 2009). In cognitive science we can take the **intentional stance** view introduced by Daniel Dennett (2013). Whereas taking a *physical stance* (in which we can predict phenomena in terms of physical laws, such as a swing) or a *design stance* (in which we can predict that an alarm clock will go off at a certain time), the intentional stance sees entities as *rational agents* having *mental notions* such as *beliefs*, *goals* and *desires*. Using this viewpoint, we assume the agent takes into account such beliefs and desires to govern its behavior. For people this is the most intuitive form of description of other people's behavior. But it is also common to use it to talk about algorithms: I can say that Google *believes* I like Lego and therefore it *desires* to feed me advertisements about it and sets a *goal* to prioritize websites referring to Lego. I can also say that Google *believes* that I *want* pizza when I enter "food" as a query since it *knows* from my profile that that is my favorite food. Dennett (2013:78): *"Anything that is usefully and voluminously predictable from the intentional stance is, by definition, an intentional system, and as we will see, many fascinating and complicated things that don't have brains or eyes or ears or hands, and hence really don't have minds, are nevertheless intentional systems."* Codementalists can now be seen as "code-mentalists", emphasizing their mental, cognitive abilities.

Unifying human behavior and AI in terms of rational agents allows me to get to the main idea of this section. Our goal is to have codementalists learn ethical behavior from humans. In the previous section I briefly explored the challenges when learning purely from observation, from scratch. However, here I have highlighted the benefits of *knowledge-rich* learning methods, where the learning process is biased (guided) by prior knowledge about the domain. What could be better *declarative*, *human* knowledge about ethical values in the documentalist domain than the previously discussed *code of ethics* from section 3? Indeed, these hold general consensus ideas on how a documentalist should behave ethically, dealing with issues such as privacy, access, and fair use of the archive or library. In addition, they are full of *intentional* descriptions, see for example: *"The Archivist should endeavour to promote access to records to the fullest extent consistent with the public interest, but he should carefully observe any proper restrictions on the use of records"*. This is clearly a bias on how the documentalist should behave and it contains a goal, a desire and several (implicit) beliefs. Codes of ethics are solid knowledge bases of the most important ethical *guidelines* for the profession, and typically they are defined to be *transparent, human-readable* and *public*, in contrast to the typical black boxes ML delivers. Using codes of ethics as a knowledge bias in adaptive documentalists that learn ethical behavior is natural, since it merely *translates* (through the rational agent connection) an *ethical code* that was designed as a bias for human behavior, and uses that as a guide or constraint.

Translating codes of ethics into knowledge biases in adaptive documentalists opens up many additional opportunities. Imagine a documentalist has been equipped with the SAA (1995) code of ethics. Learning then amounts to *finetuning* within the *boundaries* given by the code. The rule above only mentions "promote access" as a goal, but in practice the agent needs to fill specifics in from observed behavior by a documentalist. In addition, this approach supports *ethical reasoning,* for example to *resolve ethical dilemmas,* when various interests need to be balanced. Learning the value of items mentioned (such as the privacy of users, and freedom of information) can give the means to perform utilitarian (in AI called: *decision-theoretic*) calculations on what is the right behavior in the current *context.* The code of ethics puts *restrictions* on which general actions are required for particular situations, and ML fills in the details left open by the code. Users could choose[187] some particular behavioral settings themselves (e.g. Contissa et al. 2017), or such individual preferences could be learned from data. Additional knowledge that could be used as a bias are *laws* the codementalist should live by (for example privacy laws such as the DPA; see Goodall 2014 for some criticism though). My approach also allows for *enforceable* codes of ethics (as introduced in section 3), by providing the possibility to *proof* compliance with the code, as well as qualifications on how and why particular dilemmas were resolved in a certain way. Such *computational ethics* can be supported by *logics of ethics* (see Anderson and Anderson 2007 for pointers) as have been developed in AI but which, so far, often lacked statistical capabilities or ML. A final interesting aspect is that the natural distinction of the SAA (2012) between archivist *beliefs* (values) and *behavior* can easily be incorporated: beliefs are represented by a bias on the value system, whereas the behavior part biases the codementalist's action policy directly. Interestingly, this is a current topic of study in AI too. Littman et al. (2017) recently introduced a specification methodology where both reward-based specifications of tasks as well as general goal descriptions can be used.

**(6.5) Research Strategy for Ethical Codementalists: The IntERMeDIUM approach**
As a synthesis of the preceding ideas on learning, codes and intentional agents, I propose a research strategy, called **IntERMeDIUM**, with the goal to *develop ethical codementalists* in the (near) future. IntERMeDIUM refers to what Licklider wrote about (see the first section) as the link between humans and knowledge. To unite human and machine ethics, I propose to see the code of ethics as a **moral contract** between human and machine. The acronym covers the main directions on which to focus research efforts on:

- **Intentional**: The bridge between humans and machines consists of the right ontology of the (physical) world and the right level of description: beliefs, desires, intentions and goals. Codementalists should be understood as rational agents.
- **Executable**: The beliefs and desires of the codementalists need to be embedded as code that can be *executed*. Instead of asking code of ethics to be *enforceable* by punishing bad behavior after the fact, executable codes of ethics are biased by the code to ensure the right ethical behavior. Anderson and Anderson (2007:16): *"Ethics must be made computable in order to make it clear exactly how agents ought to behave in ethical dilemmas"*.
- **Reward-based**: Codementalists' ethical reasoning is based on the *human values* in the documentalist domain. The core values come from the code of ethics used to bias the agent. In addition, codementalists *finetune* their ethical behavior over time by adjusting relative values using data, feedback and experience. Experience from human documentalists is vital here, since they typically solved ethical dilemmas that arose thus far.
- **Moral**: The focus of codementalist *implementations* is on the *moral* dimension. Other skills will be developed in other, general AI research, and includes perception, mobile manipulation, reasoning with uncertainty, language interpretation and more.
- **Declarative**: All ethical bias in the codementalist is *declarative* knowledge and can be inspected at all times. Ethical inferences in specific circumstances can be *explained* in human-understandable terms. Anderson and Anderson (2007:17): *"What is critical in the explicit ethical agent distinction in our view, lies not only in who is making the ethical judgments (the machine versus the human programmer) but also in the ability to justify ethical judgments that only an explicit representation of ethical principles allows."* Ethical bias and learned ethical knowledge can be *shared* with other codementalists and general *model checking* procedures can be employed to evaluate reasoning effectiveness and correctness. Laws and regulations, such as the GDPR, can be implemented in the declarative bias to further bias the behavior of the documentalists towards legal compliance.
- **Inductive**: The codementalist is a *learning* agent. All knowledge that can not be injected as a declarative bias needs to be learned from experience or obtained from other codementalists or archivists. The codementalist's knowledge will typically not be complete, and learning should be continuing and life-long. Advanced ML needs to be implemented that allow for the codementalist asking human documentalists for advice in various ways.
- **Utilitarian**: Codementalists are *utilitarian* (collective consequentialist) moral reasoning agents. Protection of the rights of individuals is ensured by demanding that all values and decision procedures are declarative, open for inspection and transparent.

---

187 https://www.newscientist.com/article/2150330-driverless-cars-could-let-you-choose-who-survives-in-a-crash/

- **Machine**: The codementalist is a *machine*. The slow migration from human documentalists to codementalists requires that we should shift focus from humans to machines for the main operational aspects of archives, libraries and other information domains.

Based on what I have described up to here, the IntERMeDIUM is a viable, general strategy to ensure that codementalists of the future will work according to human ethical principles. Naturally many details need to be sorted out, for example on the exact translation from codes of ethics to declarative bias for AI systems, inspired by other work in e.g. medical (Anderson and Anderson (2007) and autonomous driving (Goodall, 2014 ) domains. In addition, technical challenges await concerning the specifics of the learning setting in which the codementalist will acquire its ethical values. Nevertheless, based on rapid progress in AI we can assume all such issues are *solvable* in principle. Codementalist evolution will hopefully be a smooth, progressive[188] development where the codementalist learns how to behave ethically in a human society.

What remains are several big issues and questions to start thinking about. Codementalist are not yet here, but they are coming. Key issues and questions are:
- **Appraisal - from documents to machine behavior:** The documentalist profession will change drastically. Human documentalists will become *trainers* (or, *coaches,* or *parents*) of codementalists. How will they learn these new skills and how will they be evaluated? How many trainers are needed if codementalist skills can be copied? What is the perfect *curriculum* for codementalists? Will codementalists stay a separate profession, or will they merge with other information service machines?
- **The ethics of choosing THE code of ethics**: The core of IntERMeDIUM is to inject ethical codes into machines. Out of the many possible versions, which should we pick? And who decides upon that? Documentalists, committees of experts, programmers, or more general democratic methods? For this to work, we may also need to investigate more which kinds of values hold in profession such as archivists and librarians.
- **How good a codementalist are you?** How will codementalists be evaluated? And how to evaluate *ethical performance* anyway? Maybe we can evaluate codementalists by the *"ability to participate in our culture"* (Forbus, 2016).
- **Codementalist are persons too**: depending on how codementalists and humans work and learn together in the future, important matters of *responsibilities* need to be arranged, and preferably be injected in the moral code. Depending on how sophisticated codementalists become, we may even need to think about giving them *rights*.
- **Who approves codementalists?** Depending on the impact of codementalists on daily life of people, we may need regulations concerning their *use, s*imilar to regulations concerning autonomous cars. In analogy with medicine, we may need to think about formal *approval* procedures, as a kind of "FDA for algorithms" (Tutt 2017).
- **Lifelong learning**: When documentalist practices will be changed drastically with the introduction of codementalists, ethical codes may need to be updated to reflect new social norms between humans and machines. Who decides when that time comes, and who decides what is to be changed? What to do when the time comes codementalists can do *without* human guidance, such as with the recent advances[189] in Deepmind's AlphaGo? Maybe employing codementalists also requires us to rethink ethical concepts over time (see Steinert 2014 about meta-ethics). And how to deal with codementalists going rogue (Pistono and Yampolskiy 2016)?
- **Codementalist business**: How to deal with *commercial* codementalists? Will they be different from other types? Under which circumstances can codementalists incorporate business opportunities in their ethical reasoning?

In addition, many issues are raised by the simple fact that codementalists are algorithms; thus all (ethical) concerns in this text apply. The IntERMeDIUM tries to cope with many of them by starting from a wealth of human experience contained in archival codes of ethics.

**(7) Conclusions**
In this text I have sketched a development with which the archival and library world will need to deal: the inevitable move from human documentalists to machine codementalists. I made several contributions along the way. In section 2 I described how ideas developed in the physical world can help us understand what archives and access to information are. In section 3 I surveyed ethical thinking in archives and showed that codes of ethics are a useful tool to formalize and communicate professional norms. In section 4 I have introduced a novel view on the digitalization of our society with an important role for algorithms. Section 5 surveyed the novel research area of *ethics of algorithms* and here I developed a new taxonomy based on capabilities of algorithms. Ultimately, I used all these developments to develop a strategy for obtaining digital codementalists that are capable of ethical reasoning and acting according to human norms established in professional codes of ethics. The main contribution of the text is a well-founded view on how the documentalist practice will change in our digital age, and how ethical issues that come along with that change can be handled using the same kind of digital technology using **codes of ethics as a moral contract between humans and machines.**

---
188 See for inspiration the movie Bicentennial Man (http://www.imdb.com/title/tt0182789/)
189 https://www.theverge.com/2017/10/18/16495548/deepmind-ai-go-alphago-zero-self-taught

Future research directions are plenty: both ethical and technical challenges await, and the previous section has already listed the main issues. Overall, technical sciences related to AI are required to further pursue how to encode or formalize linguistically vague and uncertain texts such as codes of ethics to executable, logical specifications. Legal and social scholars are required to analyze the status of transparent, logical and declarative specifications in relation to matters such as liability, understandability and responsibility. Furthermore, much more research is needed on the implications for the human society and humanity, but also on what it means to be human in an increasingly algorithmic, and robotic, society. The original dreams of Otlet, Bush, Wells and others of providing humans with universal access to knowledge and information are now being implemented through machines that increasingly so decide autonomously under what conditions this is happening. Keeping our human values a part of these conditions is very important.